\documentclass[conference]{IEEEtran}
\IEEEoverridecommandlockouts

\pdfoutput=1

\usepackage[switch]{lineno}  
\usepackage{cite}
\usepackage{textcomp}
\usepackage{xcolor}
\def\BibTeX{{\rm B\kern-.05em{\sc i\kern-.025em b}\kern-.08em
    T\kern-.1667em\lower.7ex\hbox{E}\kern-.125emX}}
\usepackage[utf8]{inputenc}

\usepackage{amsmath, amssymb, amsthm, amscd, amsfonts}
\usepackage{latexsym}
\usepackage{graphicx}
\usepackage{epstopdf}
\usepackage{algorithm}
\usepackage{algorithmicx}
\usepackage{algpseudocode}
\usepackage{url}
\usepackage{subfigure}
\usepackage{enumerate}
\usepackage{multirow}
\usepackage{bm}
\usepackage{array}
\usepackage[OT1]{fontenc}
\usepackage{bbding}
\usepackage{listings}
\usepackage{xspace}

\usepackage{booktabs}
\usepackage{enumitem}
\usepackage{microtype}
\usepackage{tablefootnote}
\usepackage{hyperref}

\newcommand{\method}{LightSeq2\xspace}

\newcommand{\revise}[1]{{{#1}}}

\usepackage[normalem]{ulem}
\newif\ifmodify
\modifytrue 
\ifmodify
\newcommand{\dl}[1]{\sout{#1}}
\newcommand{\guyue}[1]{\textcolor{blue}{[Guyue] {#1}}}

\newcommand{\wangxiaohui}[1]{\textcolor{green}{[Wangxiaohui] {#1}}}
\newcommand{\xiongying}[1]{\textcolor{orange}{[Xiongying] {#1}}}

\newcommand{\wy}[1]{\textcolor{pink}{[WY] {#1}}}

\else
\newcommand{\dl}[1]{}
\newcommand{\guyue}[1]{}
\newcommand{\wangxiaohui}[1]{}
\newcommand{\xiongying}[1]{}

\newcommand{\wy}[1]{}

\fi

\makeatletter
\newcommand*{\getcountref}[1]{%
\expandafter\@getcountref\csname r@#1\endcsname
}
\newcommand*{\@getcountref}[1]{%
\ifx#1\relax
0
\else
\expandafter\@car#1\@empty\@nil
\fi
}
\makeatother

\begin{document}


\title{\method: Accelerated Training for Transformer-based Models on GPUs}


\DeclareRobustCommand*{\IEEEauthorrefmark}[1]{%
  \raisebox{0pt}[0pt][0pt]{\textsuperscript{\footnotesize #1}}%
}
\author{
    \IEEEauthorblockN{
        Xiaohui Wang\IEEEauthorrefmark{1},
        Yang Wei\IEEEauthorrefmark{1},
        Ying Xiong\IEEEauthorrefmark{1},
        Guyue Huang\IEEEauthorrefmark{2},\\
        Xian Qian\IEEEauthorrefmark{1},
        Yufei Ding\IEEEauthorrefmark{2},
        Mingxuan Wang\IEEEauthorrefmark{1}, 
        Lei Li\IEEEauthorrefmark{2*}\thanks{*Partial work was done while at ByteDance.}
    }
    \vspace{0.5em}
    \IEEEauthorblockA{
        \IEEEauthorrefmark{1}ByteDance AI Lab, \IEEEauthorrefmark{2}University of California, Santa Barbara\\
        \{wangxiaohui.neo, weiyang.god, xiongying.taka, qian.xian, wangmingxuan.89\}@bytedance.com\\ 
        guyue@ucsb.edu, \{yufeiding, leili\}@cs.ucsb.edu
    }
}


\maketitle

\begin{abstract}
  Transformer-based neural models are used in many AI applications.
Training these models is expensive, as it takes huge GPU resources and long duration. 
It is challenging because typical data like sentences have variable lengths, and Transformer's computation patterns are more complex than convolutional neural networks.
Existing systems either only focus on model inference or optimization for only BERT-like encoder models.
In this paper, we present \method, a system to accelerate training for a general family of Transformer models on GPUs. 
We propose a series of GPU optimization techniques tailored to the specific computation flow and memory access patterns of Transformer models. 
\method supports many model architectures, including BERT (encoder-only), GPT (decoder-only), Transformer (encoder-decoder), and vision Transformer. 
Our experiments for a variety of models and benchmarks show that \method is consistently  faster (1.4-3.5$\times$) than previous systems on different GPUs. 
In particular, it gains 308\% training speedup compared with existing systems on a large public machine translation benchmark (WMT14 English-German).

\end{abstract}


\section{Introduction}
\label{sec:intro}

\begin{figure}[t]
\centering
\includegraphics[width=0.6\linewidth]{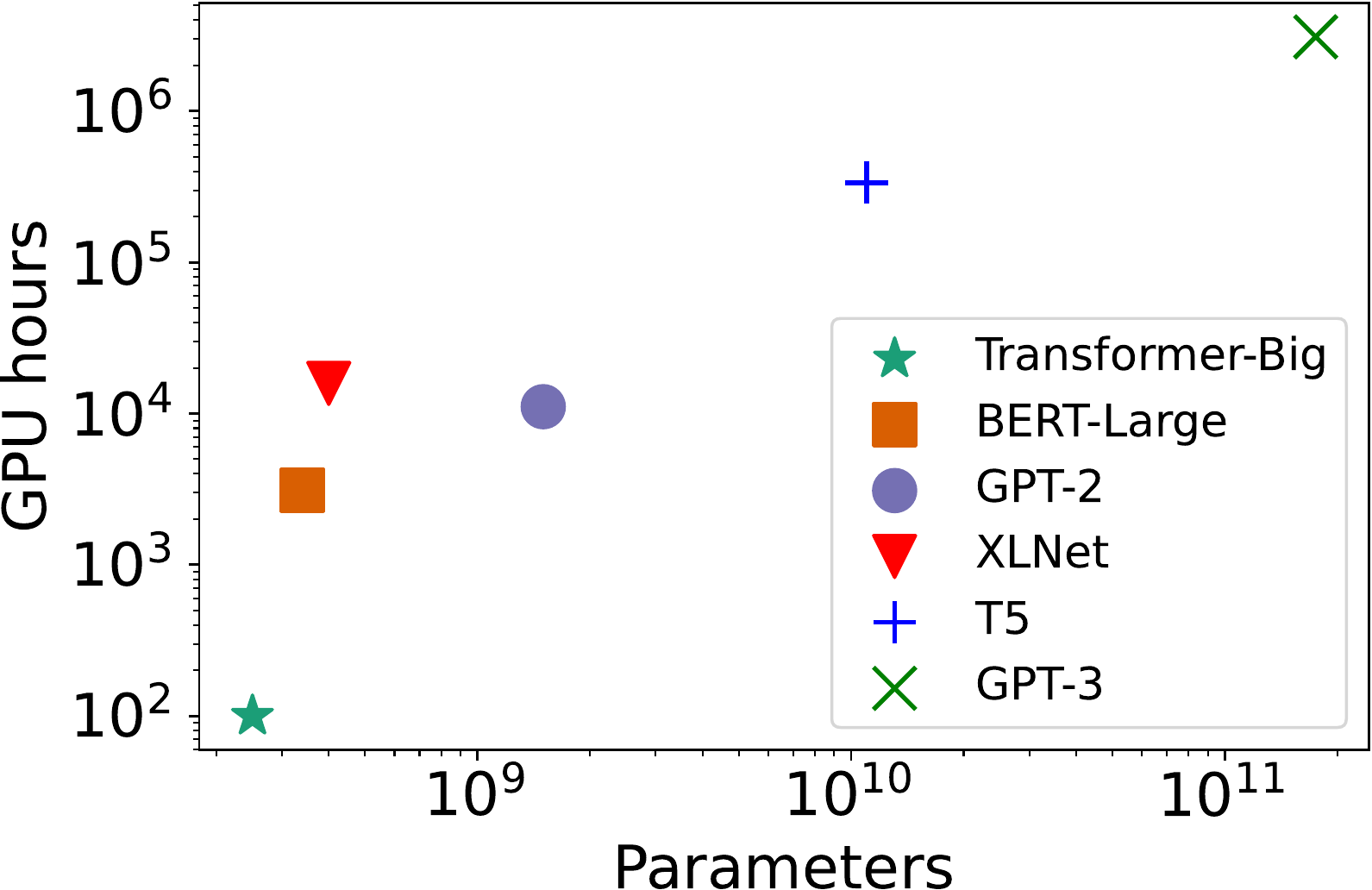}
\caption{Model size (number of parameters) and training time on Tesla V100 for popular Transformer-based models.
}
\label{fig:cost}
\end{figure}

Deep learning has been a prevailing approach to artificial intelligence (AI). 
Among various deep models, Transformers \cite{NIPS2017_3f5ee243} have become one dominant choice of model architecture in many AI tasks, including  natural language processing (NLP), computer vision (CV), and automatic speech recognition (ASR) \cite{DBLP:conf/naacl/DevlinCLT19, gpt2, DBLP:conf/nips/YangDYCSL19,DBLP:journals/corr/abs-2106-04560,gulati2020conformer}. 
Variants of Transformer have been proven to achieve state-of-the-art accuracy in text classification, question answering, machine translation, and visual object recognition tasks~\cite{DBLP:conf/naacl/DevlinCLT19,pan2021contrastive,dosovitskiy2021image}. 
Transformer models typically require large model size and training data to perform well.
For example, a GPT-3 model requires 3.1 million hours of training on modern GPUs and it costs \$4.6 million to complete a single trial~\cite{brown2020language}. 
Fig. \ref{fig:cost} shows model sizes and estimated training costs for several popular Transformer models. 
The training cost increases roughly in proportion to the number of model parameters. 
With the ever-growing model size, it becomes expensive to train them. 
Accelerating the computation for Transformers in both training and inference is critical.

\begin{table*}[ht]
\centering
\caption{Comparing systems for accelerated Transformer training. Other systems only support Transformer inference are not included. 
}
\label{tab:features}
            \resizebox{0.9\linewidth}{!}{
\begin{tabular}{c|ccccc|c|cc}
                    \toprule
                                                                \textbf{Training}   & \multicolumn{5}{c|}{\textbf{Components}}                                           & \multirow{2}{*}{\textbf{Sequence Length}}    & \multicolumn{2}{c}{\textbf{DL Frameworks}}                               \\ 
                    \textbf{Libraries} & Embedding              & Encoder                      & Decoder          & Criterion & Trainer            &  & PyTorch & TensorFlow                 \\ \midrule
                    DeepSpeed                             & {  \XSolidBrush} & { \Checkmark} & { \XSolidBrush}& {  \XSolidBrush} & {\Checkmark} & { Multiples of 16}  & { \Checkmark} & { \XSolidBrush} \\ 
                    \textbf{\method}                             & { \Checkmark} & { \Checkmark}& { \Checkmark}& { \Checkmark}& { \Checkmark} &{Arbitrary}  & { \Checkmark}& { \Checkmark}  \\ 
                    \bottomrule
                    \end{tabular}
}

\end{table*}

However, existing approaches for accelerating Transformer computation are limited to either inference-only
or encoder-only models.
LightSeq~\cite{DBLP:conf/naacl/WangXWWL21}, TurboTransformers~\cite{DBLP:conf/ppopp/FangYZZ21}, and FasterTransformer\footnote{\label{note:fastertransformer}\url{https://github.com/NVIDIA/FasterTransformer}} are recent systems targeting the serving of Transformers, but they cannot support Transformer training. 
DeepSpeed provides optimized training for Transformer~\cite{DBLP:conf/sc/RajbhandariRRH20}, but it only supports Transformer encoder layers (e.g., BERT \cite{DBLP:conf/naacl/DevlinCLT19}). 
Tasks like machine translation requires full encoder-decoder Transformer layers, criterion layers for calculating generation loss, shared embedding, etc. 
These involve more complex computation flow, as in the cross attention computation between decoder and encoder layers.
Therefore, it is nontrivial to accelerate the training for full Transformers. 

Other research on general computation acceleration for neural networks include automatic hardware-oriented compilation and quantized computation \cite{DBLP:conf/osdi/ChenMJZYSCWHCGK18,DBLP:conf/cvpr/JacobKCZTHAK18}.
However, automatic compilation only supports fixed-length input, and find it difficult to deal with a variable-length input such as natural sentences for Transformer. 
Reducing the precision by quantization is \revise{beneficial} in terms of performance but they could also lead to accuracy decrease to a certain extent, and accuracy is crucial for model training. 

Comparing to inference, there are several additional challenges to accelerate training for Transformers.  
First, in addition to forward operators in inference, a training framework needs high-performance \revise{kernels to calculate gradients of each layer during backward steps}.
\revise{Backward operators exhibit unique challenges unseen in forward operators. For instance, the sparse aggregation in Embedding's backward step requires dedicated kernel design; the complex calculation dependencies in LayerNorm's backward step hinders parallelism.} 
Second, the parameter update procedure in trainers (also known as optimizers) needs to be accelerated. 
For example, trainer takes 25\% of the time when training Transformer-Big using PyTorch on eight V100 GPUs. 
Trainers need relatively high-precision calculation for updating parameters. 
Finally, a training system needs different techniques to optimize the memory management comparing to inference systems.   
Training requires stashing all the activations of previous layers for backward usage. 
However, due to the large scale Transformer models can be, it is crucial to save GPU memory in order to train larger models.
In this paper, we focus on accelerating the training for Transformer models on modern GPUs.
We aim to provide a general system-level solution that works for all kinds of models based on Transformer, and all kinds of training algorithms such as stochastic gradient descent (SGD) and adaptive gradient methods. 

To this end, we propose \method, an efficient software library for both training and serving Transformer models. 
It provides system-level optimization without sacrificing accuracy or changing any training behavior (learning rate, convergence rate, initialization, numeric stability, etc.). 
\method includes three techniques for speedup, namely layer-specific kernels to increase GPU utilization, fine-grain mixed-precision trainer, and an improved strategy for efficient GPU memory management. \revise{Firstly, we address the GPU low utilization issue through fusing small kernels and rewriting kernels with dependencies among other parallelism enhancing techniques, based on in-depth analysis of Transformer-specific layers. Secondly, we accelerate the trainer (i.e., parameter optimizer) by employing \textit{batched} update on \textit{reduced-}precision parameters rather than many \textit{individual} updates on \textit{full-}precision parameters. Finally, we propose a memory manager that, aware of the Transformer structure, can recycle the space of tensors unused in backward pass to reduce peak memory consumption and avoid excessive allocation/release calls.} 
\method is the first to accelerate the whole process of Transformer training. 
Table \ref{tab:features} lists the differences between our proposed \method and existing accelerated Transformer training library. 
In summary,  \method enjoys the following advantages:
\begin{itemize}
    \item \textbf{Highly efficient.} \method is fast and memory efficient for training Transformers, as validated in multiple experiments. Noticeably, \method  obtains up to 308\% speedup and only requires 65\% GPU memory on eight NVIDIA Tesla A100 GPUs in WMT14 English-German machine translation task compared to PyTorch.
    \item \textbf{Supporting rich models in Transformer family.} \method provides comprehensive efficient custom operators, including embedding, encoder layer, decoder layer, and criterion. These enables BERT (encoder-only), GPT (decoder-only), full Transformer with encoder-decoder, etc. Therefore it is suitable for almost all NLP tasks such as text classification, generation, summarization, and machine translation. 
    \item \textbf{Flexible usage.} In addition to manually integrating the custom layers in model codes, the users can also use \method in popular training libraries without code modification. The library provides seamless integration with PyTorch and TensorFlow. 
\end{itemize}

The source code is available at \url{https://github.com/bytedance/lightseq}.

\section{Background}
\label{sec:background}
\begin{figure*}[ht]
\centering
\includegraphics[width=0.85\linewidth]{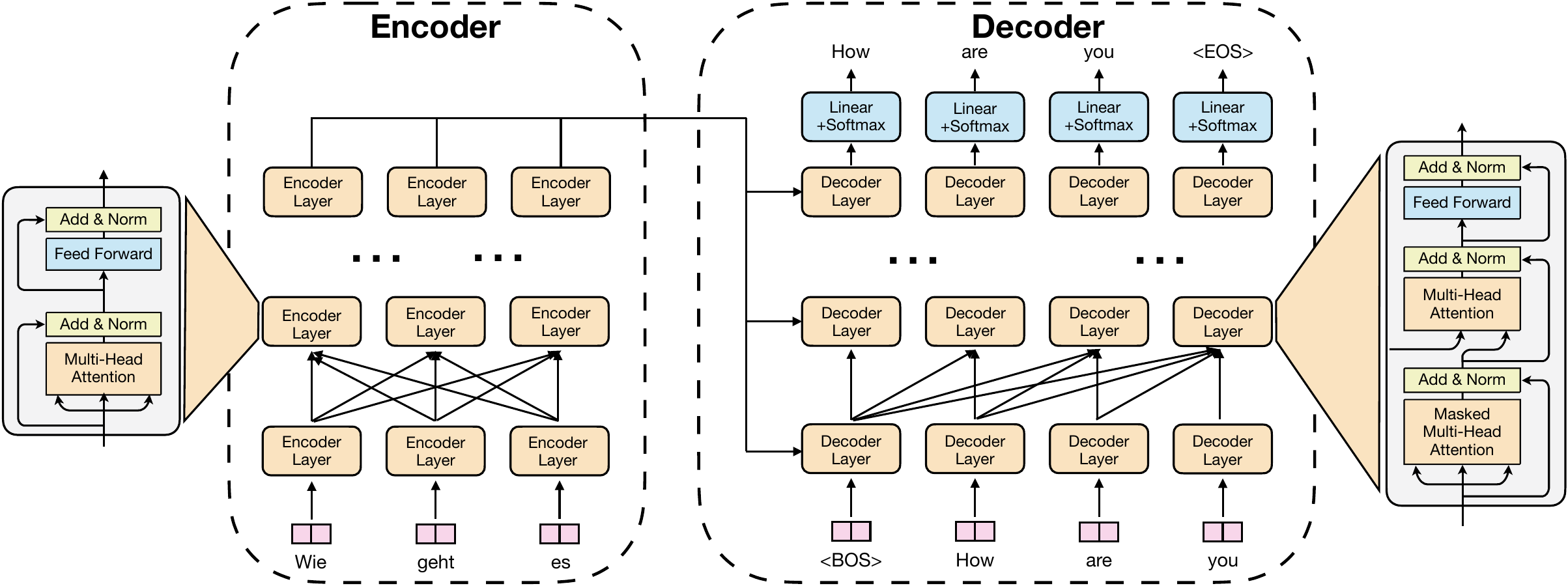}
\caption{Transformer architecture with full encoder-decoder for machine translation.}
\label{fig:transformer}
\end{figure*}

\subsection{Transformer Models}
The main idea of Transformer is using multi-head attention (MHA) to map tokens in a sequence to feature representations.
The network can be an encoder, a decoder, or an encoder-decoder architecture. 
Fig. \ref{fig:transformer} illustrates an encoder-decoder Transformer for machine translation (MT). 
Given an input sentence in the source language (e.g., German), the model first breaks the input sequence into tokens, and convert tokens into embedding vectors. 
These embeddings are processed by encoder and decoder layers. 
Both the encoder and the decoder contain multiple layers of MHA. 
Each MHA unit computes the attentional weights of one token by calculating the similarity scores to all token embeddings from the previous layer. 
The current layer's embedding is then calculated from a weighted sum of embeddings of the previous layer. 
The decoder has two differences. 
It calculates self-attention to only prefix tokens in the decoder side. 
It includes additional cross attention from decoder to encoder tokens. 
The output sequence is generated token by token where each token probability is calculated from Softmax layer. 


The Transformer-based models have variable-length intermediate tensors, which may lead to frequent allocation and release of GPU memory. 
There are several existing methods to deal with this problem. 
TurboTransformers \cite{DBLP:conf/ppopp/FangYZZ21} uses a sequence-length-aware allocator in their inference library to maximize non-dependent GPU memory sharing and memory consumption reduction. 
However, this also increases the frequencies of allocating and releasing the GPU memory.
Thus it will slow down the model inference. 
LightSeq \cite{DBLP:conf/naacl/WangXWWL21} allocates the maximal GPU memory in advance to prevent dynamic allocation and release during training.


\subsection{Model Training}
There are four stages during each iteration of data-parallel training.
\begin{enumerate}
    \item The model receives a batch of training data and then performs forward propagation, obtaining the final loss.
    \item The model performs backward propagation using the loss calculated after forward propagation, generating the gradients of all parameters.
    \item Gradients are gathered from all devices and then the averaged gradients are computed and broadcast to each device. There are two major families to complete this process, all-reduce\cite{patarasuk2009bandwidth} and Parameter Server (PS)\cite{li2014scaling}.
    \item All parameters in each device are updated using the averaged gradients. Since the initial state and gradient of the parameters on each device are the same, the parameters remain the same after one updated stage.
\end{enumerate}

\begin{figure}[t]
\centering
\includegraphics[width=0.5\linewidth]{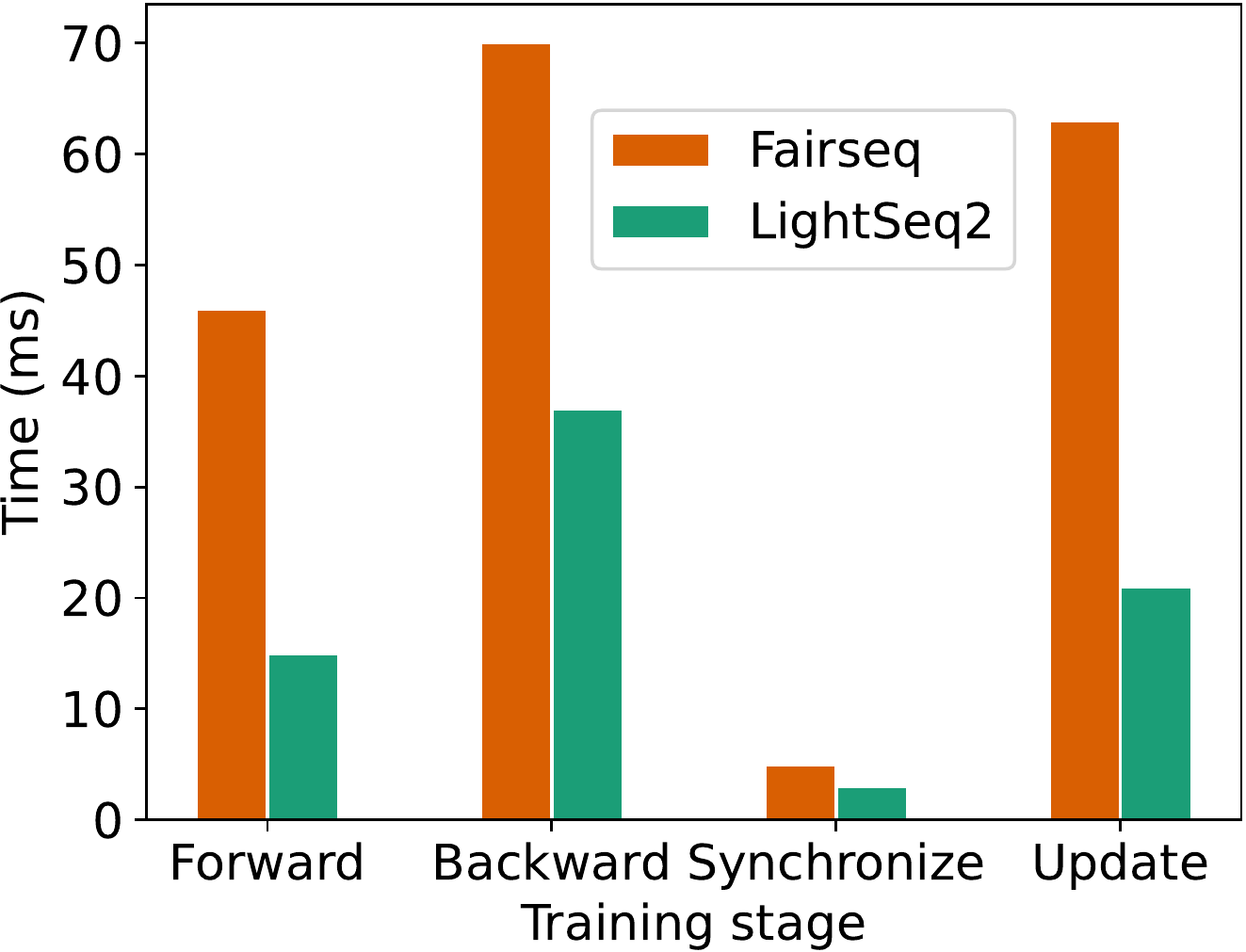}
\caption{Time cost for Fairseq and \method based on WMT-14 English-German machine translation task, using Transformer-24e24d and 8 Tesla A100 GPUs.}
\label{fig:time_dist}
\end{figure}

The bottleneck of the first and second stages is mainly in computing, which depends on fast CUDA kernels. However, the last two stages require better memory management, which will speed up the copies of parameters and gradients. Many works devote to accelerating these four stages, such as DeepSpeed, Apex, DDP \cite{DBLP:journals/pvldb/LiZVSNLPSVDC20}, etc. However, there is no work to accelerate the complete training process. Fig. \ref{fig:time_dist} is the time cost of the four training stages. It can be found that model computing and parameter updates account for a large proportion. After training with \method, the time of the three stages is greatly reduced, especially the parameter updates.

Different from model inference which only has the forward propagation stage, model training is more challenging. First, model training requires higher accuracy. Otherwise, the error will be magnified after constant parameter updates. Second, model training requires better memory management due to the need for maintaining gradients and activation checkpointing used by backward propagation. Finally, model training requires a fast trainer for parameter updates. However, for Transformer models, the training needs not deal with incremental length in auto regressive decoding, which is simpler than inference.

\section{Related Work}
\label{sec:related}

Many approaches have been proposed to speed up the computation for deep neural models, which can be divided into two categories, algorithm-specific and algorithm-agnostic.

Algorithm-specific methods accelerate the training process by improving the model architectures \cite{DBLP:conf/nips/VyasKF20,DBLP:journals/corr/abs-2102-12702,DBLP:journals/corr/abs-2006-04768, DBLP:conf/iclr/FanGJ20,DBLP:conf/nips/ZhangH20,peng2021random,choromanski2021rethinking}, training strategies \cite{DBLP:conf/iclr/LiuJHCLG020, pmlr-v97-gong19a, DBLP:conf/emnlp/LiWLJDXWZ20}, optimization algorithms \cite{DBLP:journals/corr/abs-1904-00962, DBLP:journals/corr/abs-2006-00719}, data precision \cite{DBLP:conf/cvpr/ZhangLZLHZGGDZC20,NEURIPS2019_65fc9fb4,NEURIPS2018_335d3d1c}, and others \cite{DBLP:conf/aaai/LiWLDXZZ21, DBLP:conf/icml/LiWSLKKG20}. 
Vyas et al.~\cite{DBLP:conf/nips/VyasKF20} use clustered attention to group queries into clusters and compute attention just for the centroids, resulting in a linear complexity model. 
Gong et al.\cite{pmlr-v97-gong19a} propose a stacking algorithm to transfer knowledge from a shallow model to a deep model, then applies stacking progressively to accelerate BERT training. Yang et al.\cite{DBLP:journals/corr/abs-1904-00962} propose ADAHESSIAN, a second-order stochastic optimization algorithm that dynamically incorporates the curvature of the loss function via adaptively estimates of the Hessian matrix.
These techniques can speed up the model training to a certain extent but may affect the model structure and effect, so the universality is not good.

For algorithm-agnostic optimization, 
Apex\footnote{\url{https://github.com/NVIDIA/apex}} developed many commonly used GPU kernels using C++ and CUDA, including \texttt{LayerNorm}, \texttt{Softmax} and Adam optimizer. 
It supports automatic mixed precision computation similar to earlier quantization idea~\cite{DBLP:conf/cvpr/JacobKCZTHAK18}   and distributed training. 
Unlike previous works that change the training behavior, the engineering level optimization strictly follows the training algorithm and has no impact on anything other than speed. 
Our \method further enhances the performance of trainer with memory-efficient mixed precision computation without sacrificing the accuracy.

DeepSpeed \cite{rasley2020deepspeed} integrates these small kernels into Transformer encoders, which boosts the training throughput on a single GPU and scales well on multiple GPUs. However, DeepSpeed has several limitations which hinder its usage in NLP, CV, and ASR tasks, especially in sequence generation tasks. First, DeepSpeed only optimizes Transformer encoder, thus is not suitable for tasks requiring decoding modules (e.g., machine translation). Second, DeepSpeed does not optimize the other module like embedding and criterion, which prevents it achieving higher performance. Third, DeepSpeed requires that the input length be an integer multiple of 16 due to the implementations of some kernels, which introduces unnecessary padding and computation. In contrast, \method supports the arbitrary shape of inputs. Fourth, DeepSpeed does not support TensorFlow, which is also widely used in practice.

LightSeq (version 1.2)~\cite{DBLP:conf/naacl/WangXWWL21}, TurboTransformers (version 0.5)~\cite{DBLP:conf/ppopp/FangYZZ21}, and FasterTransformer (version 4.0)\footnotemark[1] are recent systems targeting the serving of Transformers. All systems exploit manually written CUDA kernels for accelerated forward computation of layers in a Transformer. 
They also improve the serving throughput by enhanced batch decoding strategies on GPUs. 
However, neither supports Transformer training since there are additional backward computation which is more complex than the forward pass. 

TVM \cite{DBLP:conf/osdi/ChenMJZYSCWHCGK18} is a compiler that searches for candidate optimizations in the neural network according to specific patterns and automatically merge operators. 
It exposes graph-level and operator-level optimizations to provide performance portability to deep learning models. 
However, it is difficult to apply to Transformer-based models due to the variable input lengths.

\section{The \method System}
\label{sec:approach}
This section will introduce the four techniques used in \method in detail, including computational graph optimizations, dependent reduction rewriting, accelerated mixed-precision update for trainer, and dangling-tensor aware memory manager.

\subsection{Computational Graph Optimizations}\label{sec:approch-graph-optimize}

Deep learning frameworks usually use computational graphs to represent the programs. Nodes represent operations like addition and multiplication, and edges represent tensor data flowing between operations.

We replace straightforward fine-grained GPU kernel functions in PyTorch or TensorFlow implementations with coarse-grain fused ones. Our customized operators avoid high overhead introduced by a mass of kernel function launches and GPU memory I/O for intermediate results during forward/backward propagation.

\subsubsection{Transformer Layers}
To all types of Transformers, we design fused kernel operators for both encoder and decoder layers.

The computational graph of a Transformer has two types of operations. One involves GEMM, including linear transformation and scaled dot product. The other is non-GEMM, such as \texttt{Dropout}, \texttt{LayerNorm}, and \texttt{Softmax}. To ensure the flexibility to cover various variants of the transformer-based models, we focus on fusing non-GEMM kernels and directly use the GEMM implementations from cuBLAS.The non-GEMM operations can be further grouped into two categories. One is element-wise operation (e.g., \texttt{Dropout}, \texttt{ReLU}, \texttt{Reshape} and bias adding), and the other is reduction operations (e.g. \texttt{LayerNorm} and \texttt{Softmax}). Element-wise operations enable explicit parallelism and multi-kernel fusion. Reduction operations require careful synchronization between threads. 

\begin{figure}[t]
\centering
\includegraphics[width=\linewidth]{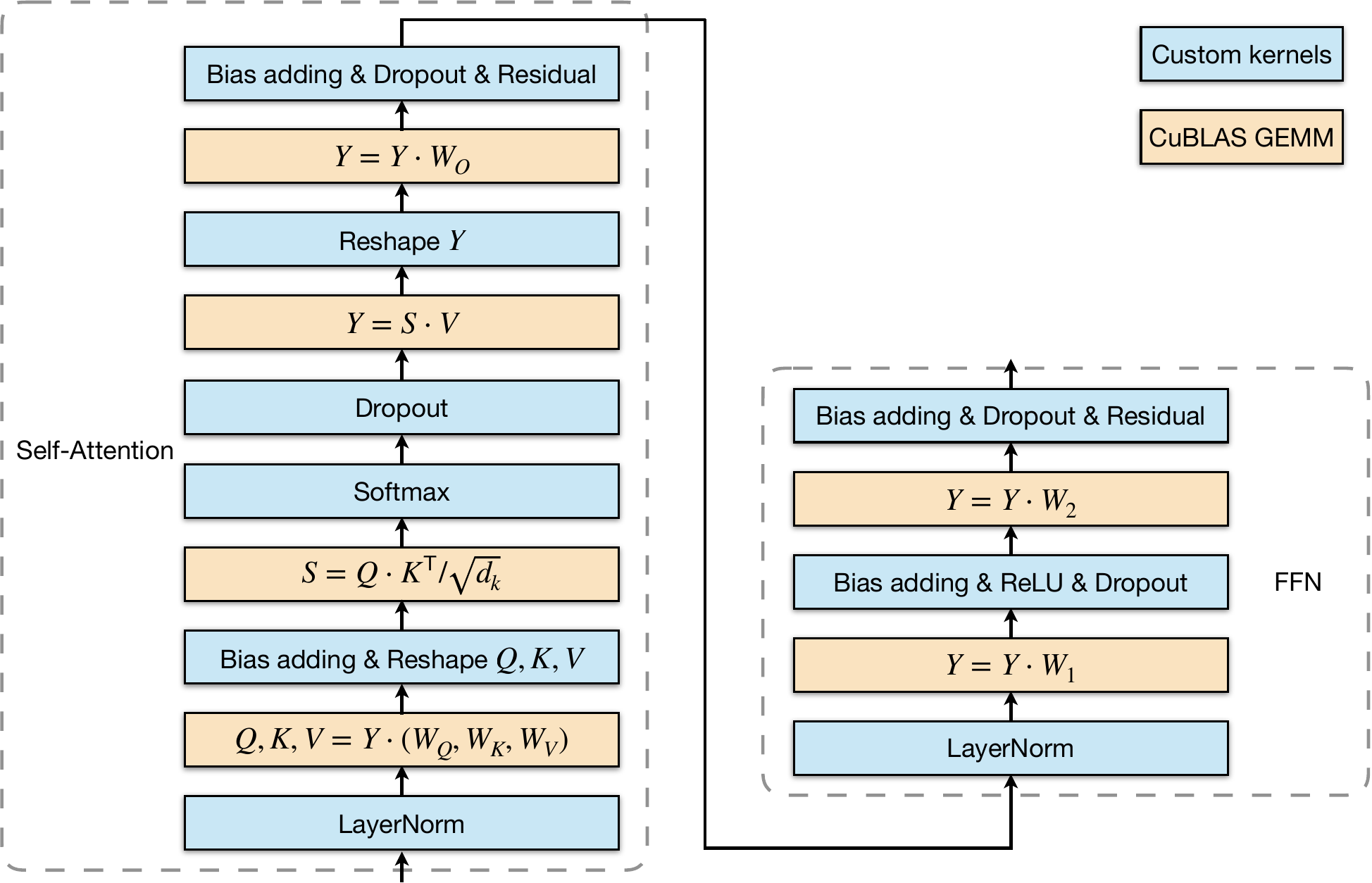}
\caption{\method's optimized computational graph for Transformer with pre-\texttt{LayerNorm}.}
\label{fig:forward}
\end{figure}

Fig. \ref{fig:forward} shows optimized computational graph for Transformer. The yellow boxes represent GEMM kernels and blue boxes represent custom non-GEMM kernels. Adjacent fine-grained element-wise kernels are fused into one coarse-grained kernel, resulting in fewer kernel launches and intermediate results. For example, the last kernel of the self-attention layer implements bias adding, dropout, and residual kernels with only one kernel launch.

\subsubsection{Embedding Layer}


The embedding layer is widely used in most deep learning tasks to obtain the distributed representation of one word or an image patch. Given a token embedding lookup table $\mathbf E$ and positional embedding lookup table $\mathbf P$, we can get the representation $\mathbf{y}^{(w, p)}$ of one token with index $w$ and position $p$:
\[
\mathbf{y}^{(w, p)} = \mathrm{Dropout}(s \cdot \mathbf{E}_{w} + \mathbf{P}_{p}),
\]
where $s$ is the embedding scale.

Considering an input sentence $x$ with length $l$, let $\mathbf m$ denotes the \texttt{Dropout} mask generated in the forward propagation. We can efficiently compute the gradient of token $w$ in the embedding table:
\[
\nabla \mathbf{E} = s \cdot \sum_{0 \le i < l, \mathbf{x}_{i} = w}{\mathbf{m}^{(i)} \odot \nabla \mathbf{y}^{(\mathbf{x}_{i}, i)}},
\]
where $\odot$ represents an element-wise product. \revise{We use $\nabla x$ to denote $\frac{\partial{\mathcal{L}}}{\partial{x}}$, where $\mathcal L$ is loss from criterion layer, and $x$ is any intermediate layer output.} This means summing up all gradients of tokens $w$ in the different positions in the sentence, which can be implemented in parallel by \texttt{atomicAdd} operation in CUDA to avoid interference from other threads.

\subsubsection{Criterion Layer} 
The criterion layer is used to compute the loss between the model output and the ground truth. The loss is cross entropy for machine translation. Let $\mathbf h$, a vector of length $V$, denote the output of the decoder for one token, where $V$ is the vocabulary size, and $\mathbf y$ denote the one-hot vector of length $V$ representing the ground truth. The prevalent cross-entropy loss with label smoothing can be formulated as 
\[
\mathcal{L}(\mathbf p, \mathbf q) = -\sum_i \mathbf{p}_i \log(\mathbf{q}_i),
\]
where 
$\mathbf p = (1 - \alpha)  \mathbf y + \frac{\alpha}{V}\cdot \mathbf 1$ and  
$\mathbf q = \mathrm{Softmax}(\mathbf h)$. 
$0\le \alpha\le 1$ is the smoothing parameter, and $\sum_i \mathbf q_i=1$.

By plugging $\mathbf p$ and $\mathbf q$ into the loss function and calculating the partial derivatives of $\mathbf{h}_i$, we can get the gradient of decoder output as follows.

\revise{First, the gradient of \texttt{Softmax} function is:
\[
\frac{\partial{\mathbf{q}_i}}{\partial{\mathbf{h}_j}} = \left\{
\begin{array}{ll}
   -\mathbf{q}_i\mathbf{q}_j  & i \neq j \\
    \mathbf{q}_i(1-\mathbf{q}_i) & i=j
\end{array}  \right.\nonumber,
\]
We derive the gradient to decoder output in two situations. If $i$ is equal to the ground truth token index $k$, the gradient as:
\[
\begin{aligned}
    \nabla \mathbf{h}_i  &= \frac{\partial \mathcal{L}}{\partial \mathbf{h}_i} = -\frac{\alpha}{V}\sum_{j \neq k}{\frac{1}{\mathbf{q}_j}\cdot \frac{\partial{\mathbf{q}_j}}{\partial{\mathbf{h}_k}}} - (1 - \alpha + \frac{\alpha}{V}) \cdot \frac{1}{\mathbf{q}_k} \cdot \frac{\partial{\mathbf{q}_k}}{\partial{\mathbf{h}_k}} \\
    &= \mathbf{q}_k -\frac{\alpha}{V}  - 1 + \alpha
\end{aligned}.
\]
The gradient for other slots are:
\[
\begin{aligned}
    \nabla \mathbf{h}_i &=\frac{\partial \mathcal{L}}{\partial \mathbf{h}_i} = -\frac{\alpha}{V}\sum_{j \neq k}{\frac{1}{\mathbf{q}_j}\cdot \frac{\partial{\mathbf{q}_j}}{\partial{\mathbf{h}_i}}} - (1 - \alpha + \frac{\alpha}{V}) \cdot \frac{1}{\mathbf{q}_k} \cdot \frac{\partial{\mathbf{q}_k}}{\partial{\mathbf{h}_i}} \\
    &= -\frac{\alpha}{V}\sum_{j \neq k, j \neq i}{\frac{1}{\mathbf{q}_j}\cdot \frac{\partial{\mathbf{q}_j}}{\partial{\mathbf{h}_i}}} -\frac{\alpha}{V}\cdot{\frac{1}{\mathbf{q}_i}\cdot \frac{\partial{\mathbf{q}_i}}{\partial{\mathbf{h}_i}}} \\
    &\quad - (1 - \alpha + \frac{\alpha}{V}) \cdot \frac{1}{\mathbf{q}_k} \cdot \frac{\partial{\mathbf{q}_k}}{\partial{\mathbf{h}_i}} 
    = \mathbf{q}_i -\frac{\alpha}{V}
\end{aligned}.
\]
Therefore the final gradient to the decoder output is:
}
\[
\nabla \mathbf{h}_i = \left\{
\begin{array}{ll}
   \mathbf{q}_i -\frac{\alpha}{V}  - 1 + \alpha  & \mathrm{if\ token\ }i\mathrm{\ is\ the\ ground\ truth} \\
    \mathbf{q}_i  -\frac{\alpha}{V} & \mathrm{otherwise}
\end{array}  \right.\nonumber,
\]
which is an element-wise kernel of $\mathbf q$ and can be executed in parallel. Our \texttt{Softmax} operation will be introduced later. But we do not directly calculate it. Instead, we calculate logarithm of \texttt{Softmax} by adding additional logarithmic operations and bias adding for forward and backward calculation.

\begin{figure}[t]
\centering
\subfigure[Original.] {
    \label{fig:cross-attn-1}
    \includegraphics[width=0.46\linewidth]{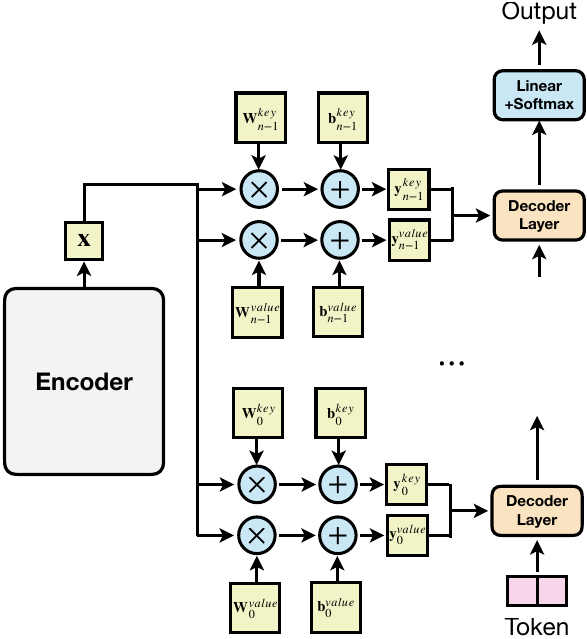}
}
\subfigure[\method.] {
    \label{fig:cross-attn-2}
    \includegraphics[width=0.46\linewidth]{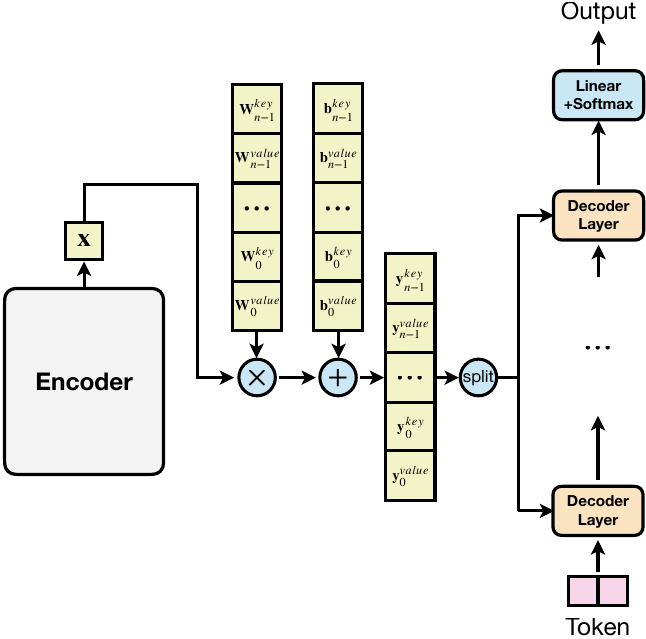}
}
\caption{The computation of cross attention in \method.}
\label{fig:cross-attn}
\end{figure}

\subsubsection{Layer-batched cross Attention}
As shown in Fig. \ref{fig:cross-attn-1}, considering the model has $n$ decoder layers, the context (vectors like keys and values) in cross attention of decoder layer is obtained by calculating the linear transformation of the encoder output:
\[
\begin{aligned}
    \textbf{y}^{key}_i &= {\mathbf{W}}^{key}_{i} \cdot \textbf{x} + {\mathbf{b}}^{key}_{i} \\
    \textbf{y}^{value}_i &= {\mathbf{W}}^{value}_{i} \cdot \textbf{x} + {\mathbf{b}}^{value}_{i}
\end{aligned}
\]
where $\textbf{y}^{key}_i$ and $\textbf{y}^{value}_i$ are the contexts in cross attention of $i$-th decoder layer. ${\mathbf{W}}^{key}_{i}$ and ${\mathbf{W}}^{value}_{i}$ are the parameter matrices. ${\mathbf{b}}^{key}_{i}$ and ${\mathbf{b}}^{value}_{i}$ are the bias vectors. $\mathbf{x}$ is the output of the encoder.
As shown in Fig. \ref{fig:cross-attn-2}, we merged the computation of the context of cross attention to reduce the launch of kernels and improve the concurrency of matrix multiplication:
\[
[{\textbf{y}^{key}}; {\textbf{y}^{value}}] = [{\mathbf{W}}^{key}; {{\mathbf{W}}^{value}}] \cdot \mathbf{x}  + [{\mathbf{b}}^{key}; {\mathbf{b}}^{value}],
\]
where
\[
\begin{aligned}
     {{\mathbf{W}}^{key}} &= [ {{\mathbf{W}}^{key}_0}; {{\mathbf{W}}^{key}_1}; \ldots; {{\mathbf{W}}^{key}_{n-1}} ] \\
     {{\mathbf{W}}^{value}} &= [ {{\mathbf{W}}^{value}_0}; {{\mathbf{W}}^{value}_1}; \ldots; {{\mathbf{W}}^{value}_{n-1}} ]
\end{aligned}.
\]
Splitting $[{\textbf{y}^{key}}; {\textbf{y}^{value}}]$ once will obtain context of all decoder layers in forward propagation. For backward propagation, \method computes the gradient of loss with respect to $\mathbf{x}$ (encoder output) after finishing the backward propagation of $0$-th decoder layer.

\subsection{Dependent Reduction Rewriting}
\method optimizes two batch reduction operations traditionally taking a long time in the training, including \texttt{LayerNorm} and \texttt{Softmax}.

\noindent\textbf{LayerNorm} normalizes the inputs $\mathbf{x}$ using 
\[
\mathbf{y}_i = \mathbf{w}_i \cdot \frac{\mathbf{x}_i - \mu(\mathbf{x})}{\sigma(\mathbf{x})} + \mathbf{b}_i,
\]
where $\mu(\mathbf{x})$ and $\sigma(\mathbf{x})$ stand for the mean and standard variance of $\mathbf{x}$ respectively. Both are batch reduction operations. While warp-level parallelism provided by CUDA allows inter-thread communication for batch reduction over $\mathbf{x}$, we cannot merge these two reductions directly. Because calculating variance requires the mean.
To speed up the forward pass, \method adopts the following formula for reduction of variance:
\[
\sigma(\mathbf{x}) =\sqrt{ \mu(\mathbf{x}^2) - \mu(\mathbf{x})^2}.
\]
In two reduction operations, both means of $\mathbf{x}$ and $\mathbf{x}^2$ are computed in parallel.

\begin{figure}[t]
\centering
\includegraphics[width=\linewidth]{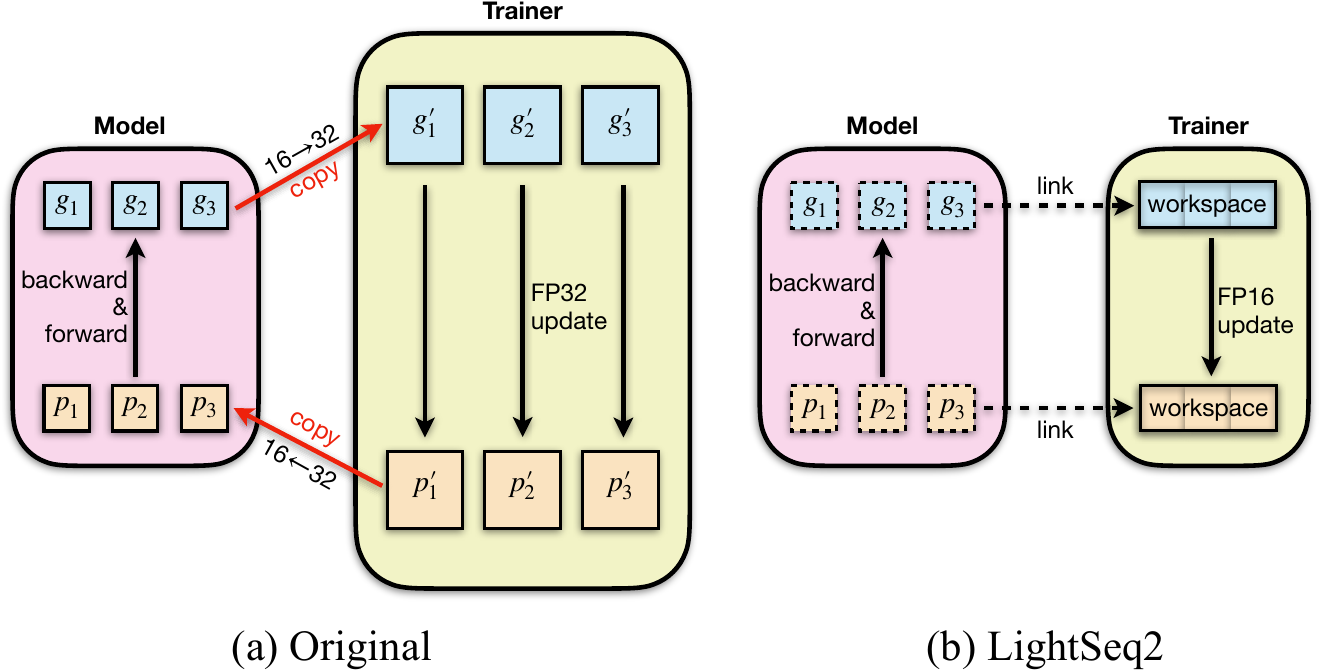}
\caption{Mixed-precision trainer in \method.Orange and blue blocks represent the model parameters and gradients respectively. Dotted blocks represent symbolic tensor link which have no actual memory storage. \method eliminates FP32 copies of parameters/gradients and directly updates the FP16 workspace with on-the-fly conversion.}
\label{fig:opt}
\end{figure}

\revise{
To calculate the gradient, first let $\hat{\mathbf{x}}_i = \frac{\mathbf{x}_i - \mu(\mathbf{x})}{\sigma(\mathbf{x})}$. Then the gradient of $\mathbf{x}_i$ is:
\[
\nabla \mathbf{x}_i = \sum_{j}{\nabla{\mathbf{y}_j} \cdot \mathbf{w}_j \cdot \frac{\partial{\hat{\mathbf{x}}_j}}{\partial{\mathbf{x}_i}}}.
\]
\revise{The derivation process of normalization partial differential} $\frac{\partial{\hat{\mathbf{x}}_j}}{\partial{\mathbf{x}_i}}$ \revise{is not introduced in detail here for its popularity. Thus the final gradient is:}
\[
\nabla \mathbf{x}_i = \frac{\mathbf{w}_i \nabla{\mathbf{y}_i}}{\sigma(\mathbf{x})}  - \frac{1}{m\sigma(\mathbf{x})}\left( \sum_{j}{\nabla{\mathbf{y}_j} \mathbf{w}_j} + \hat{\mathbf{x}}_i\sum_{j}{\nabla{\mathbf{y}_j} \mathbf{w}_j}\hat{\mathbf{x}}_j  \right),
\]
where $m$ is the dimension of $\mathbf x$, $\nabla \mathbf{x}_i$ and $\nabla \mathbf{y}_i$ are the gradients of the $i$-th elements of input and output tensors respectively.
}

\method rearrange the gradient calculation formula to enable parallel computation:
\[
\nabla \mathbf{x}_i = \frac{\mathbf{w}_i \nabla \mathbf{y}_i}{\sigma(\mathbf x)} + \alpha \cdot \sum_j \mathbf{w}_j \nabla \mathbf{y}_j  + \beta \cdot \sum_j  \mathbf{w}_j  \nabla \mathbf{y}_j \mathbf{x}_j,
\]
where $\alpha$ and $\beta$ are coefficients that can be solved in parallel:
\[
\begin{aligned}
    \alpha &= \frac{[\mathbf{x}_i - \mu(\mathbf{x})]\mu(\mathbf{x}) - \sigma(\mathbf{x})^2}{m \sigma(\mathbf{x})^3} 
    \quad\quad \beta = \frac{\mu(\mathbf{x}) - \mathbf{x}_i}{m \sigma(\mathbf{x})^3}
\end{aligned}.
\]
The two batch reductions $\sum_j \mathbf{w}_j \nabla \mathbf{y}_j $ and $\sum_j \nabla  \mathbf{w}_j \mathbf{y}_j  \mathbf{x}_j$ can be executed in parallel.

As \texttt{LayerNorm} is sensitive to the precision of floating points, \method stores half-precision float point (FP16) for parameters and casts them to single-precision float point (FP32) during computation to avoid additional I/O cost.

\noindent\textbf{Softmax} Attention in Transformer needs \texttt{Softmax}. The forward process of \texttt{Softmax} can be expressed as
$\mathbf{y}_i = \frac{\exp(\mathbf{x}_i)}{\sum_j \exp(\mathbf{x}_j)}$.

For numerical stability, especially for mixed-precision training, it takes three steps to avoid overflow:
\begin{enumerate}
    \item Find the maximal element of $\mathbf{x}$, denoted as $x'$.
    \item Deduct $x'$ from each element in $\mathbf{x}$ so that the exponential never overflows, and then calculate the partition function  $Z = \sum_j \exp(\mathbf{x}_j- x')$.
    \item Calculate \texttt{Softmax} using $\mathbf{y}_i = \exp(\mathbf{x}_i- x') / Z$.
\end{enumerate}
Both step 1 and 2 are reduction operations. The number of reductions and reduction dimension are quite diverse in the attention \texttt{Softmax}. For example, in common NLP tasks, the reduction dimension ranges from a few to thousands and the number of reductions ranges from thousands to millions. To address this challenge, we implement multiple templates suitable for diverse shapes and let key parameters (e.g., number of blocks, warps per block, reduce times per block) tunable. \method runs templates and searches for parameters before training to determine the optimal configuration for each shape.


\begin{figure}[t]
\centering
\subfigure[Symbolic target link (Python).] {
    \label{fig:code1}
    \includegraphics[width=0.46\linewidth]{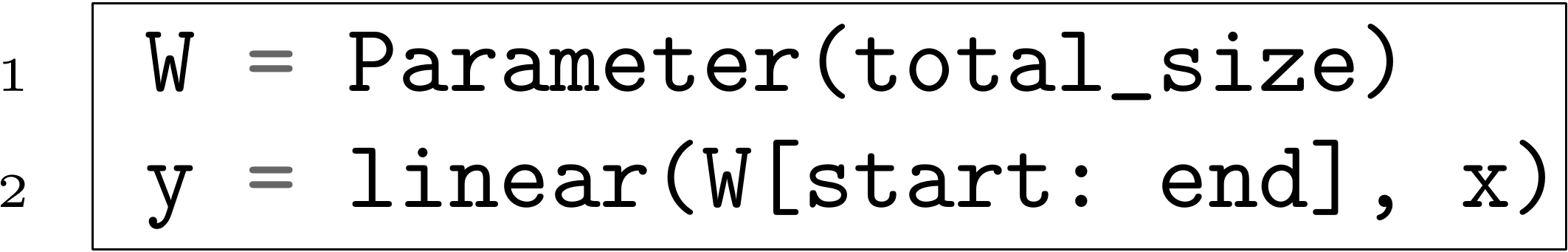}
}
\subfigure[On-the-fly (GPU kernel).] {
    \label{fig:code2}
    \includegraphics[width=0.46\linewidth]{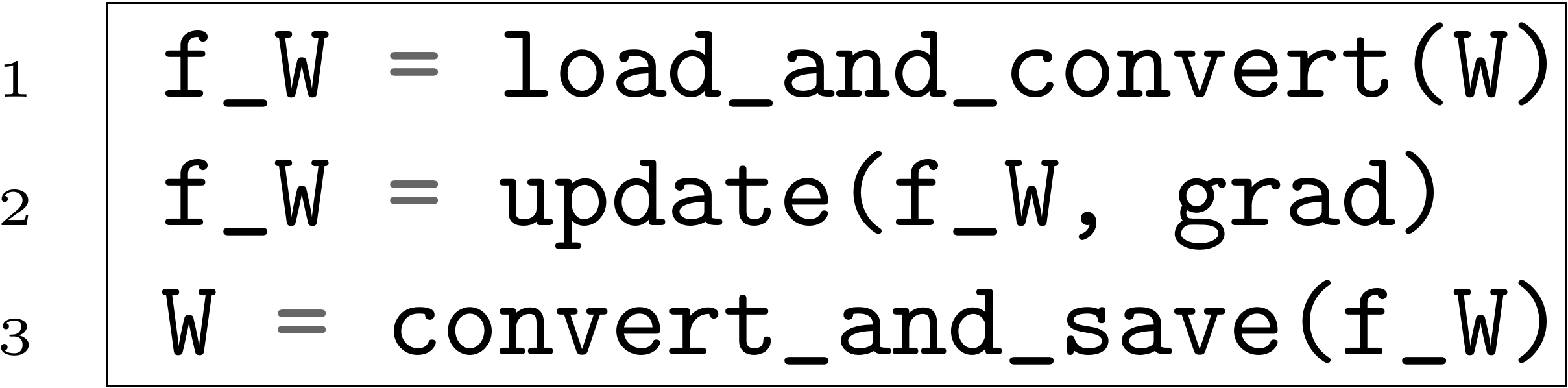}
}
\caption{\revise{A simple implementation of the two key technologies in \method trainer. (a) Symbolic target link is implemented with Python. It defines all parameters into a continuous workspace (Line 1), and then uses corresponding memory blocks to compute (Line 2). (b) On-the-fly mechanism is implemented with GPU kernel. It first loads the only one parameter and converts it to float32 precision (Line 1). Then it updates the float32 parameters (Line 2). Finally, it converts the parameters back to float16 precision and saves it into original memory (Line 3).}}
\label{fig:trainer-code}
\end{figure}

\subsection{Accelerated Mixed-Precision Update for Trainer}\label{sec:approch-adam}
Trainer include numerical update algorithm for parameters (e.g. Adam). In mixed precision training \cite{mixed-precision-training}, parameters and gradients are in FP16 during forward and backward propagation. Since the update values, the product of learning rates and gradients, are often tiny, the trainer need to maintain FP32 copies of parameters and gradients to ensure accuracy. A straightforward system copies each piece of gradients/parameters in the model to/from its FP32 partner in one training step (Fig. \ref{fig:opt}(a)). The trainer kernel will load the FP32 gradient to update the FP32 parameters. This mechanism has two disadvantages:
\begin{enumerate}
    \item Numerous pieces of gradients/parameters lead to multiple fast-returning GPU kernels like copying and updating, which reduce GPU utilization.
    \item Redundant memory footprints are caused by the FP32 copy of gradients/parameters.
\end{enumerate}

\begin{figure}[t]
\centering
\includegraphics[width=\linewidth]{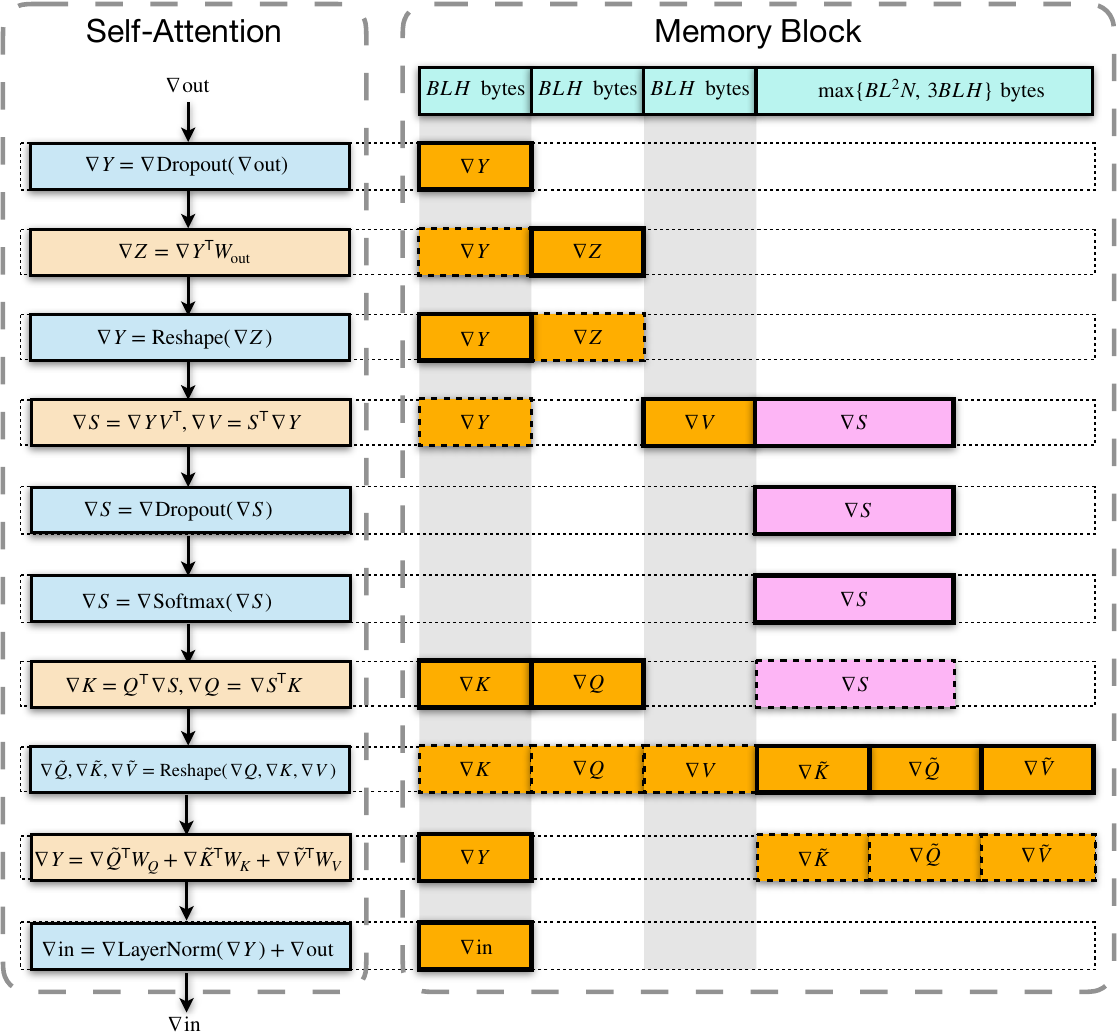}
\caption{\method's memory manager reuses pre-allocated memory as much as possible. \revise{Tensors in the same column on the right side have no dependencies, hence sharing the same pre-allocated memory block. So \method only needs to allocate the memory with the maximal size (the width of the memory block on the right side) once before training starts.}}
\label{fig:mem}
\end{figure}

Our \method alleviates them by the symbolic tensor linking and the on-the-fly conversion. During the initialization of the trainer, \method copies all pieces of parameters/gradients into one tensor called workspace orderly (Fig. \ref{fig:opt}(b)). Then we reset and link them as fragments of workspace. During each training step, \method only executes the trainer kernel once to update the workspace, which prevents launching huge amount of fragmented GPU kernels on every piece of parameters/gradients.

Our trainer kernel loads the FP16 parameters/gradient from workspace to register and converts it on-the-fly to FP32. Then the parameters on register will be updated as usual. Finally, the parameters will be converted on-the-fly to FP16 and saved to workspace. Accessing memory with FP16 instead of FP32 reduces the data movement by half and avoids the FP32 copies of parameters and gradients.

The cooperation between the symbolic tensor linking and the on-the-fly conversion leads to both memory savings and latency reduction without reducing accuracy. The experimental result on Transformer-big model shows that the proposed trainer reduces its memory usage by 2 GB and its runtime by 54.9\% compared to the Fairseq trainer with high kernel fusion from Apex. \revise{Fig. \ref{fig:trainer-code} is a simple implementation of the two technologies.}

\subsection{Dangling-Tensor Aware Memory Manager}
In practice, training with a large batch contributes to fast convergence and higher accuracy. However, large batch training requires multiple GPU, gradient accumulation, or memory offload due to the memory limit. 
 
\method reduces the memory footprint by compacting memory with fewer allocation-and-releases at no extra cost. \method divides the GPU memory into permanent memory with a fixed size to store parameters and gradients, and temporary memory with variable sizes to store intermediate tensors. To avoid the frequent allocation of temporary memory, we scan the training set and estimate the upper bound of the capacity. Thus temporary memory was allocated once with the maximal size before training starts, and it is reused for different batches and released after training finishes.

\begin{figure}[t]
\centering
\includegraphics[width=0.9\linewidth]{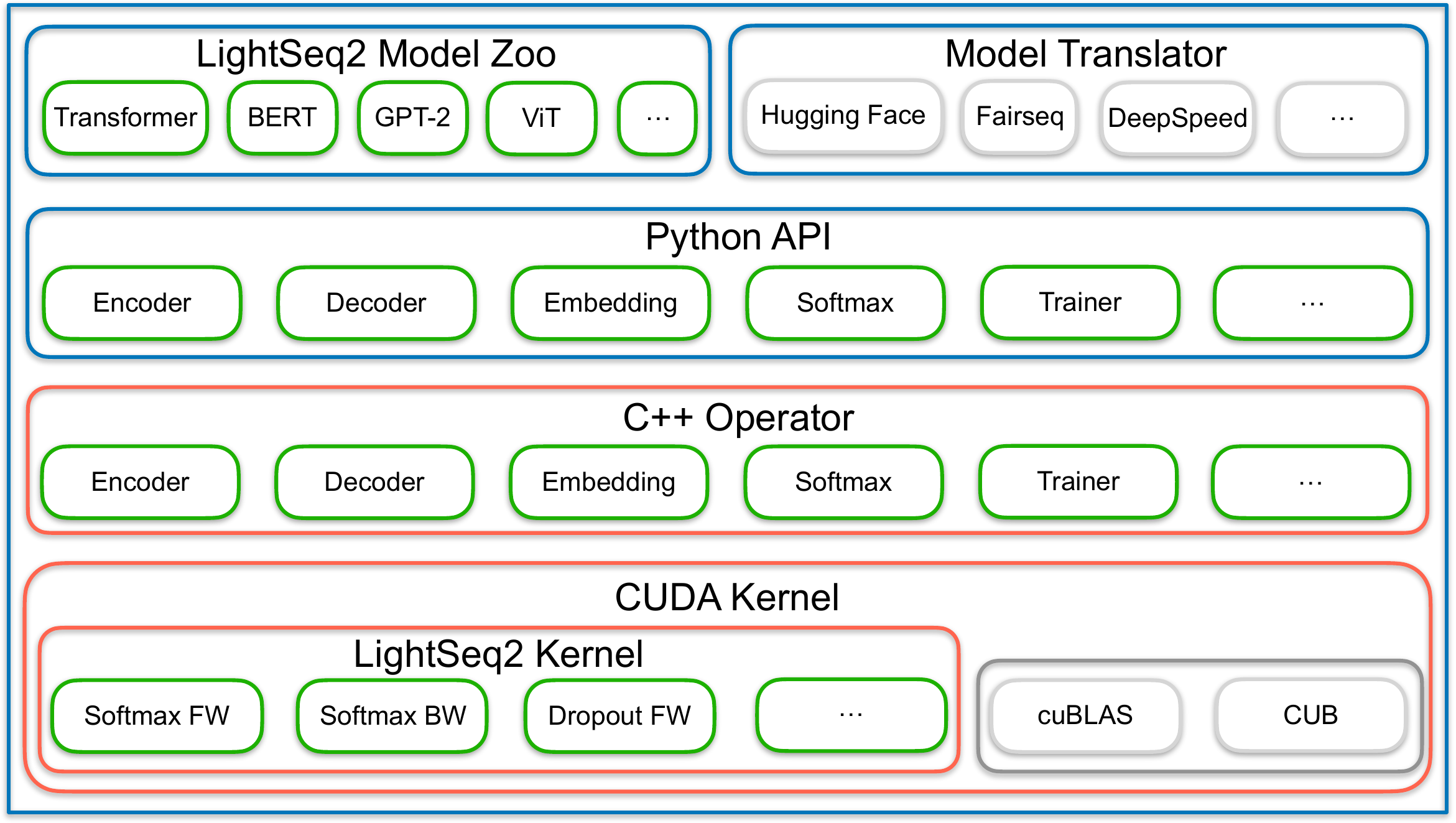}
\caption{\method software architecture, where FW represents forward and BW represents backward. Python API supports both PyTorch and TensorFlow. Blue represents external interfaces}
\label{fig:framework}
\end{figure}

\begin{figure*}[t]
\centering
\subfigure[6e6d on V100.] {
    \label{fig:fp16_pt_v100_6e6d_8gpu_apex-ls}
    \includegraphics[width=0.239\linewidth]{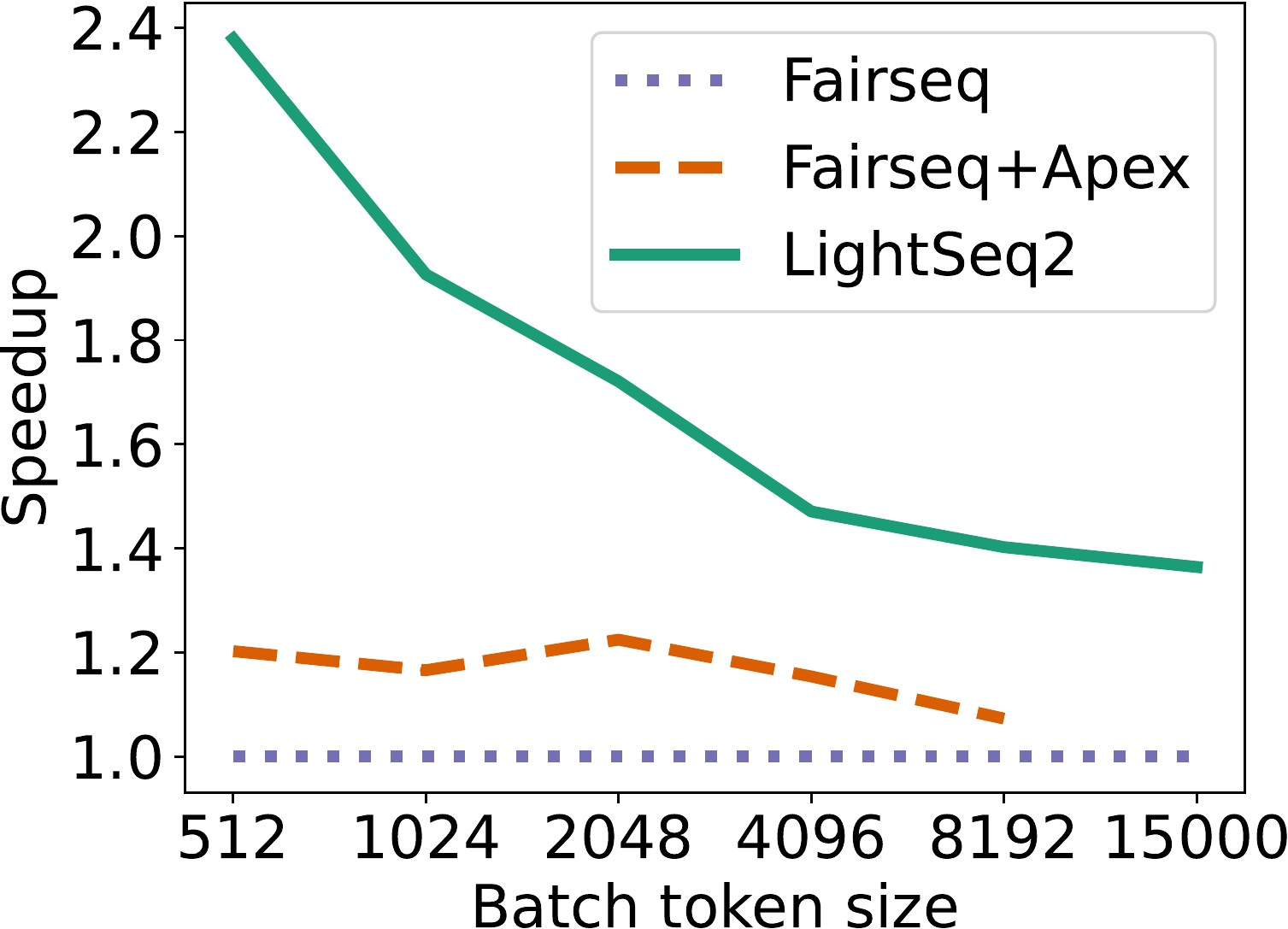}
}
\subfigure[12e12d on V100.] {
    \label{fig:fp16_pt_v100_12e12d_8gpu_apex-ls}
    \includegraphics[width=0.239\linewidth]{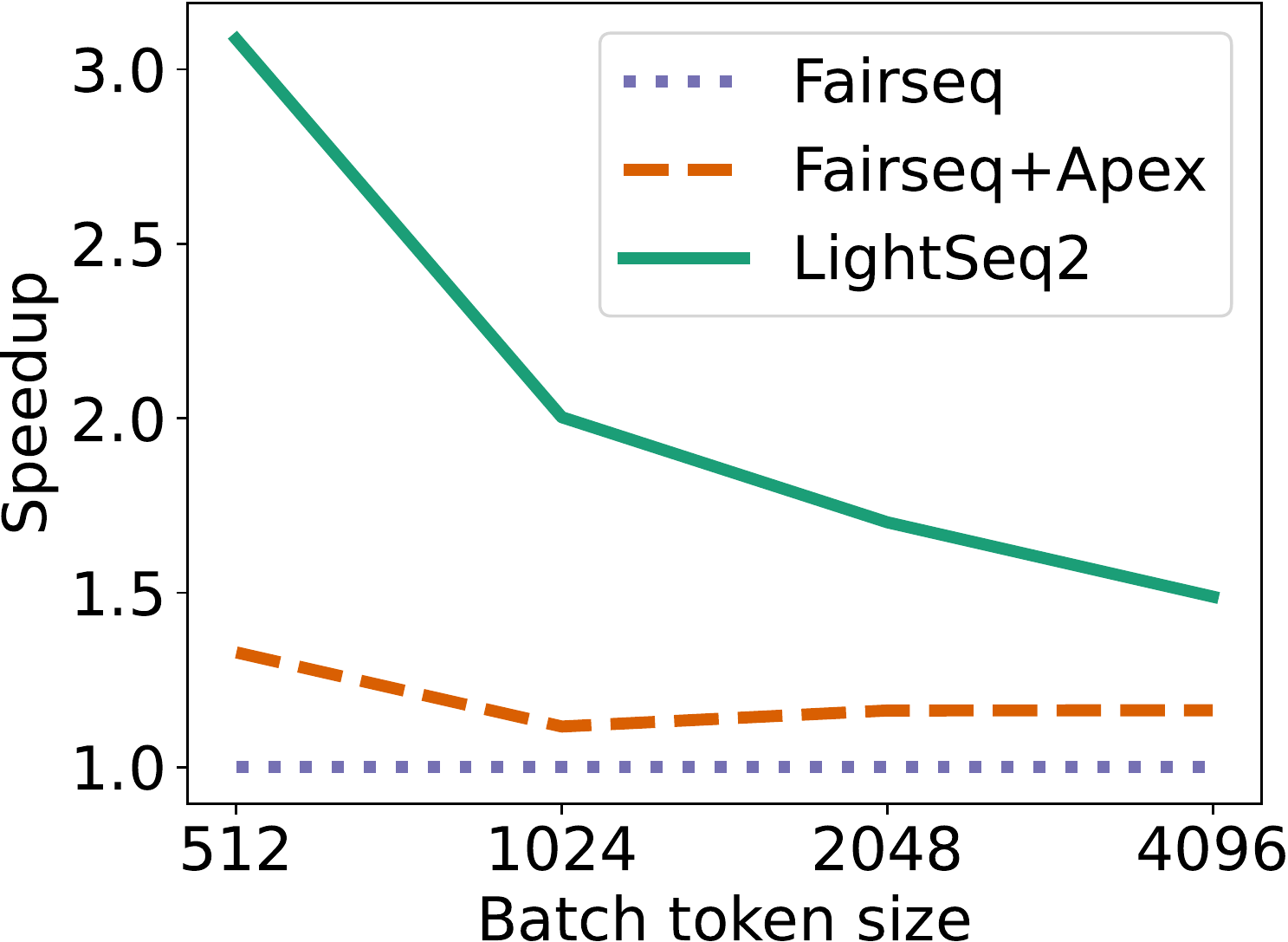}
}
\subfigure[24e24d on V100.] {
    \label{fig:fp16_pt_v100_24e24d_8gpu_apex-ls}
    \includegraphics[width=0.239\linewidth]{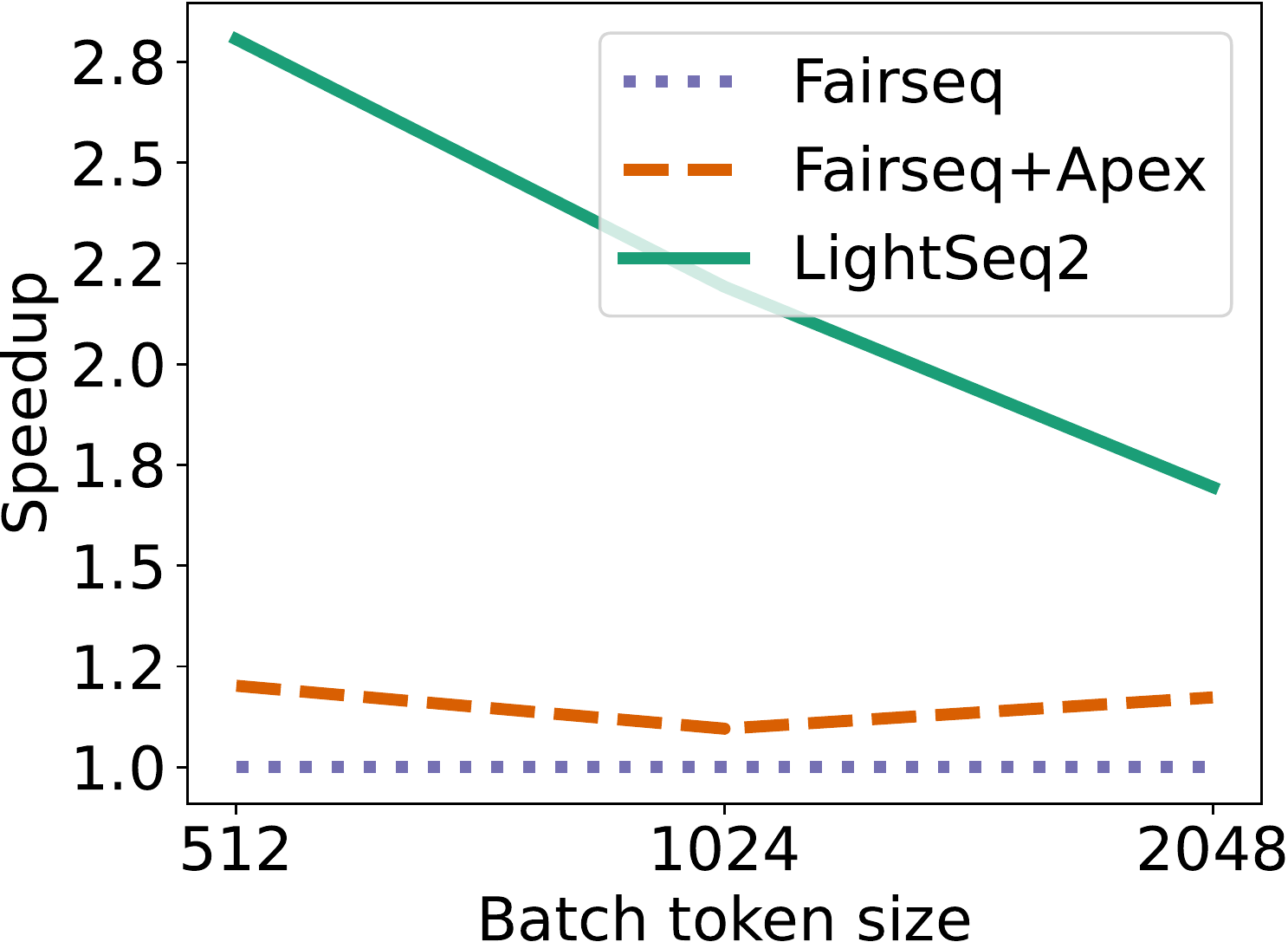}
}
\subfigure[6e6d on A100.] {
    \label{fig:fp16_pt_a100_6e6d_8gpu_apex-ls}
    \includegraphics[width=0.239\linewidth]{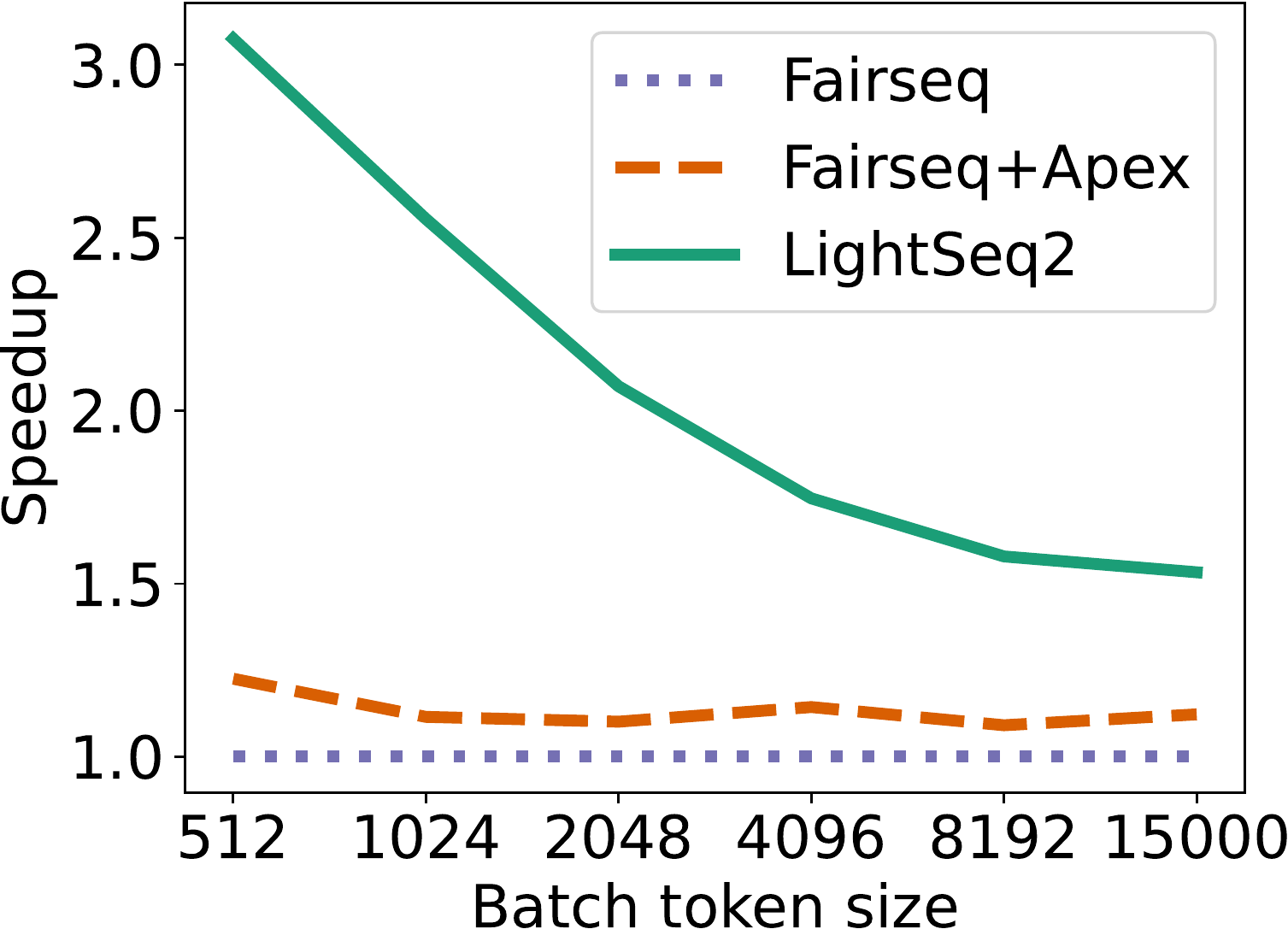}
}
\subfigure[12e12d on A100.] {
    \label{fig:fp16_pt_a100_12e12d_8gpu_apex-ls}
    \includegraphics[width=0.239\linewidth]{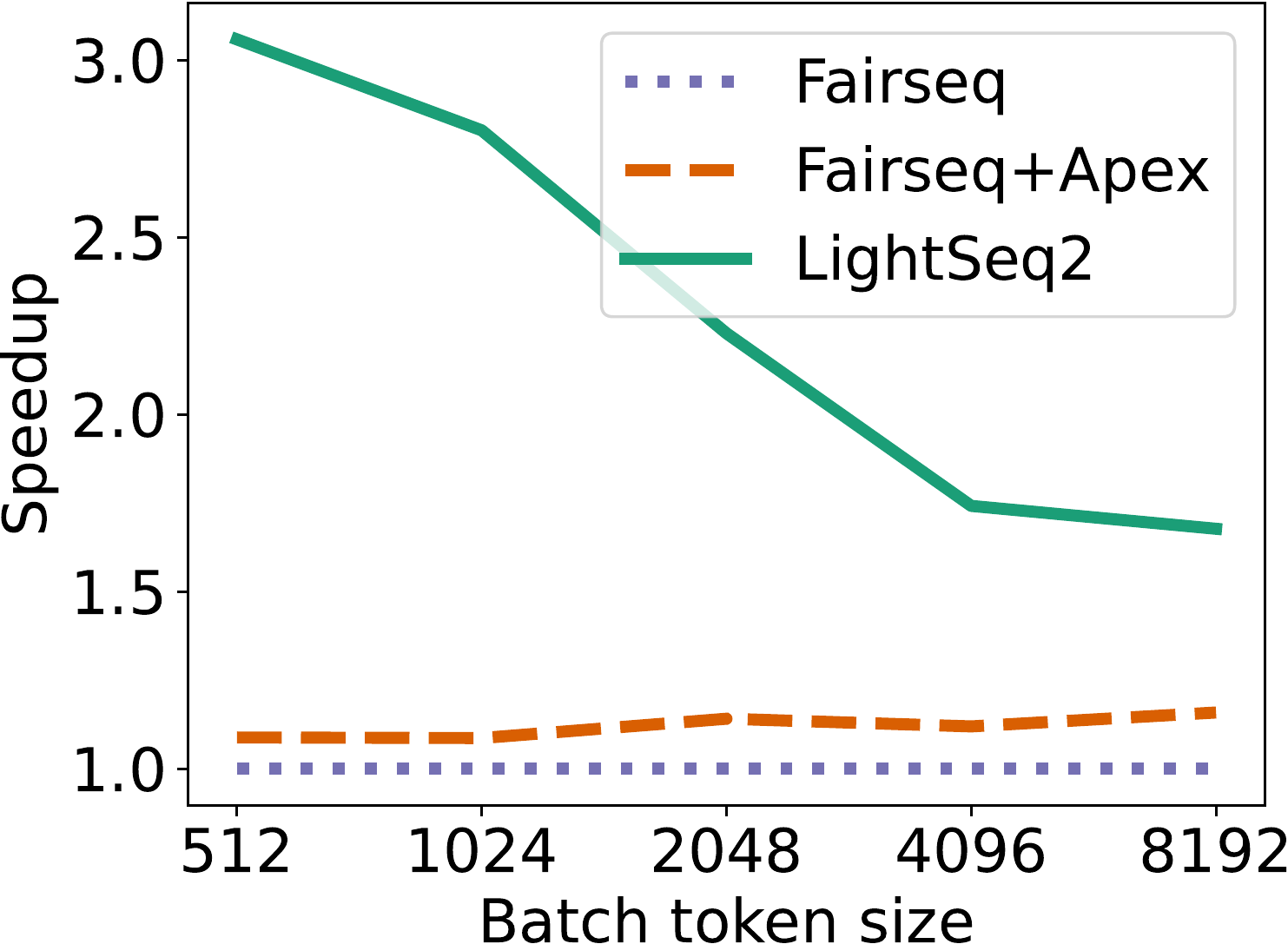}
}
\subfigure[24e24d on A100.] {
    \label{fig:fp16_pt_a100_24e24d_8gpu_apex-ls}
    \includegraphics[width=0.239\linewidth]{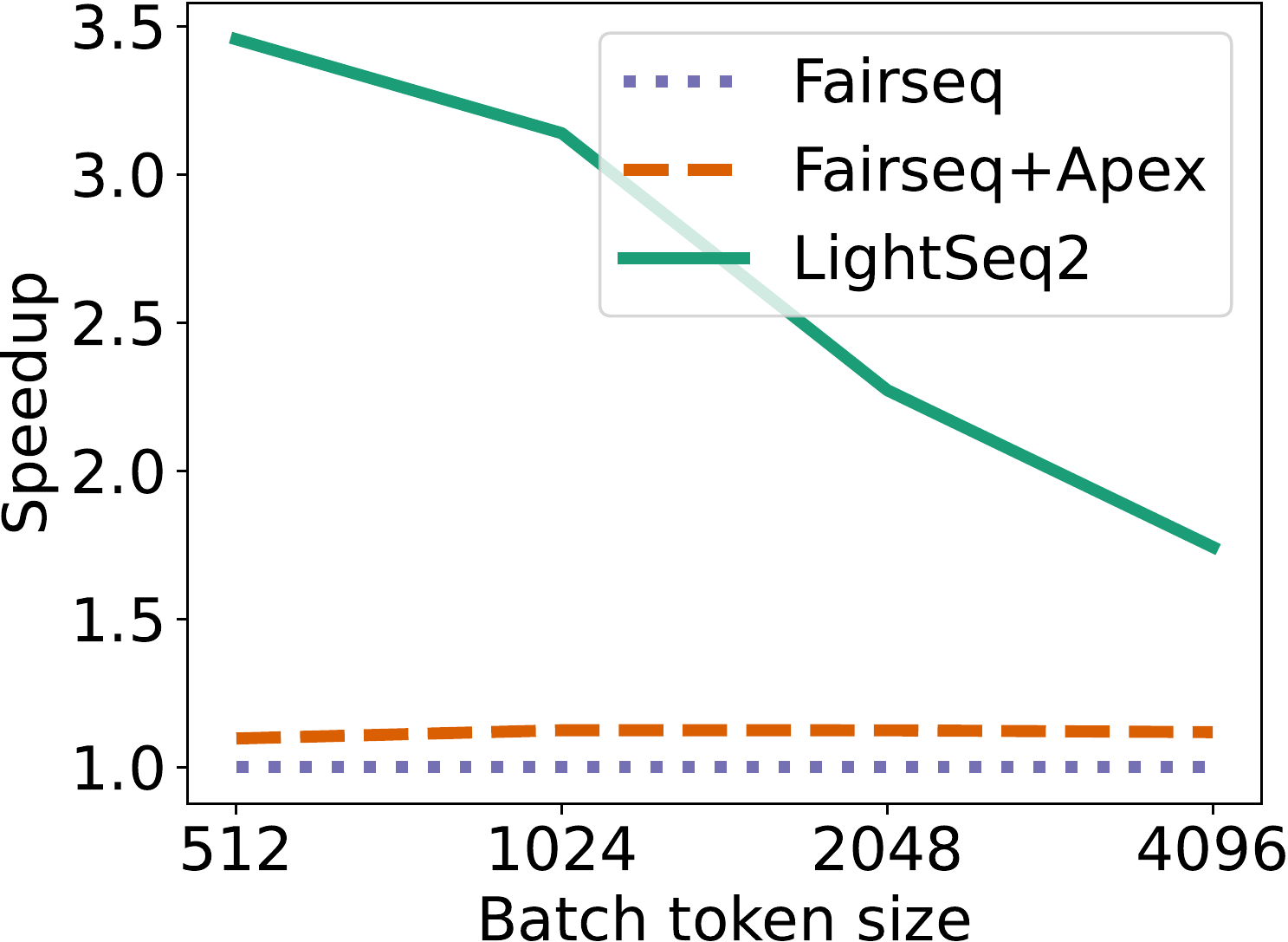}
}
\caption{Training speed comparison on machine translation using different numbers of Transformer layers on both V100 and A100 GPUs. 6e6d denotes 6 encoder layers and 6 decoder layers.}
\label{fig:fp16_pt_8gpu_apex-ls}
\end{figure*}

Fig. \ref{fig:mem} shows \method's temporary memory cost in the self-attention backward process. The left side describes the steps of backpropagation. Each row on the right side lists the memory occupancy of temporary tensors in one step. Tensors in the same column reuse the same memory block. The sizes of the orange and purple tensors are $BLH$ and $BL^2N$. The tensors in the dashed memory blocks are not updated in that step, and those in the solid blocks are updated. We can see that only $3BHL$ (the first three blocks) $ + \max\{3BHL, BL^2N\}$ (the last block) bytes of memory are required, where $B, H, L, N$ denote the batch size, hidden dimension, max sequence length and the number of heads respectively. In contrast, without using the shared memory block strategy, a total of $9BLH + BL^2N$ bytes of memory will be required.

\section{Usability Design}
\label{sec:usability}
We design \method to be modular, usable and interoperable.
\subsection{Software Architecture}
The software architecture of \method consists of five modules, as shown in Fig. \ref{fig:framework}. The lower two are internal and the upper three are exposed to users.

The CUDA kernel module implements efficient CUDA kernel operations, such as Softmax forward, Softmax backward, etc. These kernels are optimized for various input shapes. \method directly calls the external cuBLAS library for other kernels like matrix multiplication (GEMM).

The C++ operator module encapsulates common operators in C++ using above CUDA kernels to includes both forward and backward procedures for encoders and decoders. In addition, \method also utilizes GPU memory reusing technology to save GPU memory.

The Python API module provides many convenient APIs in Python by encapsulating C++ operators. With these APIs, Transformer model training can be implemented through just a few lines of codes. The Python API supports models trained with both PyTorch and TensorFlow.

The model zoo module implements many common models using Python APIs above. \method provides a model translator to export models developed in other high-level frameworks (e.g., Fairseq and Hugging Face) to our own operators.

\subsection{Interoperability}
\method is deeply integrated with mainstream training codebases, and users can use \method for accelerated training without modifying the code. For example, in Fairseq, users only need to modify the startup command to \method-train and specify the \method Transformer module. In addition, the original model in PyTorch or TensorFlow and \method model can be easily converted to each other to support accelerated fine-tuning.

\subsection{Example Usage}
\label{sec:implementation:example_usage}

\begin{figure}[t]
\centering
\includegraphics[width=0.9\linewidth]{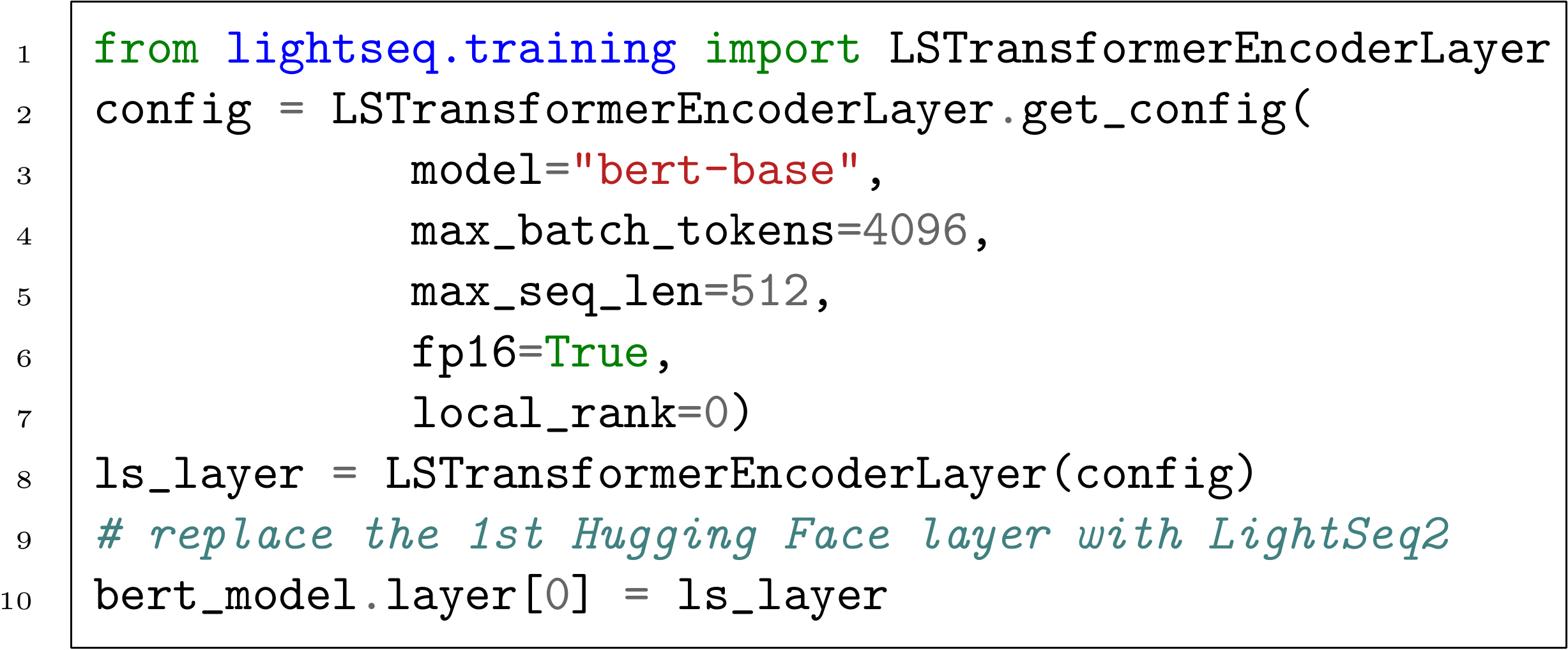}
\caption{\revise{An example of Hugging Face BERT acceleration with \method. After creating the \method encoder layers, we only need to replace the Hugging Face layers with them, and then train the model as normal.}}
\label{fig:code}
\end{figure}
We provide both C++ and Python APIs for users. 
\revise{Fig. \ref{fig:code} is a code snippet of accelerating Hugging Face BERT training with \method.} 
\method provides a rich amount of built-in, size-reconfigurable network architectures. Line 2-7 are used to provide the configuration information of the encoder layer.

\section{Experiments}
\label{sec:experiment}
We investigate the following questions in expertments.
\begin{enumerate}
    \item Does \method obtain consistent \textbf{overall speedup} for a variety of models across different GPUs?
    \item How much benefit does each optimization technique bring?
    \item How much is the memory saving in \method?
    \revise{\item How does \method scale with respect to model sizes and the number of GPUs?}
\end{enumerate}

\begin{table}[t]
\centering
\caption{Models in experiments (implemented in both \method and baselines).}
\label{tab:setup}
\begin{tabular}{cclcc}
\toprule
\textbf{Models} & \textbf{Encoder} & \textbf{Decoder}  & \textbf{Baseline} \\ \midrule
Transformer & {\Checkmark} & {\Checkmark} & Fairseq + Apex \\ \midrule
ViT & {\Checkmark} & {\XSolidBrush} & Hugging Face \\ \midrule
BERT & {\Checkmark} & {\XSolidBrush} & DeepSpeed \\ \midrule
GPT-2 & {\XSolidBrush} & {\Checkmark} & Hugging Face \\ \bottomrule
\end{tabular}
\end{table}

\subsection{Overall Speedup}\label{sec:eval:end2end}
\subsubsection{Setup}
\textbf{Benchmarks}: We include four tasks in the experiments (Table \ref{tab:setup}): machine translation using WMT14 English-German data \cite{bojar-EtAl:2014:W14-33}, image classification using CIFAR-10 data \cite{Krizhevsky09learningmultiple}, text classification using GLUE data \cite{DBLP:conf/iclr/WangSMHLB19}, and language modeling using WikiText data \cite{DBLP:conf/iclr/MerityX0S17}. The models are Transformer (with full encoder-decoder), ViT, BERT and GPT-2.
We evaluate each network architecture with different sizes and layer configurations to accommodate the need of a wide range of use cases.

\textbf{Platforms}: We run experiments using NVIDIA Tesla V100 and Tesla A100.

\textbf{Baselines}: We choose the state-of-the-art implementations for the benchmark tasks as the baselines: \textbf{Fairseq} \cite{DBLP:conf/naacl/OttEBFGNGA19}, \textbf{Hugging Face}\footnote{\url{https://github.com/huggingface/transformers}}, \textbf{Apex}\footnote{\url{https://github.com/NVIDIA/apex}}, and \textbf{DeepSpeed}\footnote{\url{https://github.com/microsoft/DeepSpeed}}. As shown in Table \ref{tab:setup}, we use the out-of-the-box end-to-end implementation provided by these baselines and tune the setup based on our run-time environment. \revise{To measure the speed, we use words per second compared with Fairseq and samples per second compared with Hugging Face.}

\begin{figure}[t]
\centering
\subfigure[ViT-B-32.] {
    \label{fig:vit_b_32}
    \includegraphics[width=0.42\linewidth]{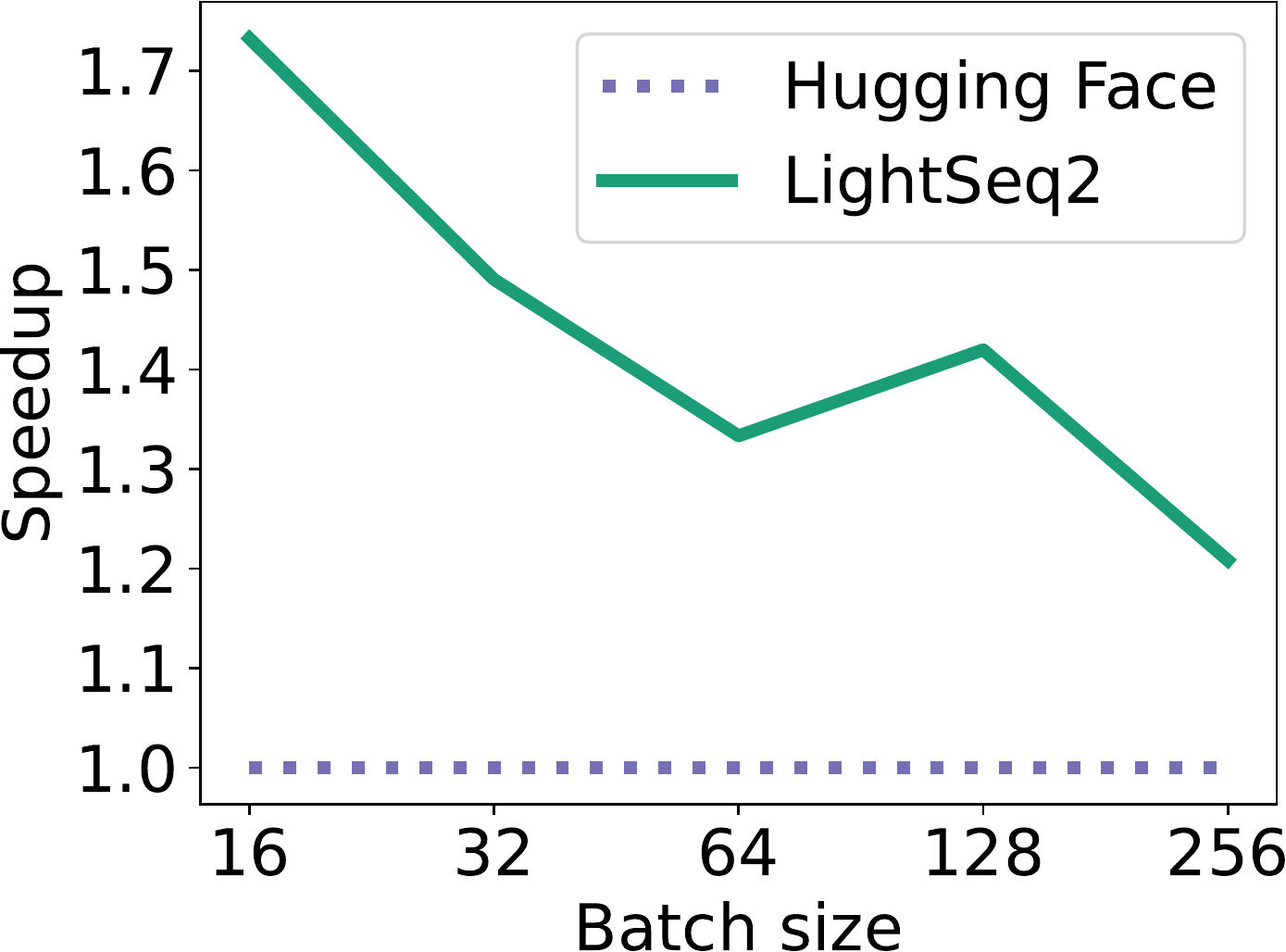}
}
\subfigure[ViT-L-32.] {
    \label{fig:vit_l_32}
    \includegraphics[width=0.42\linewidth]{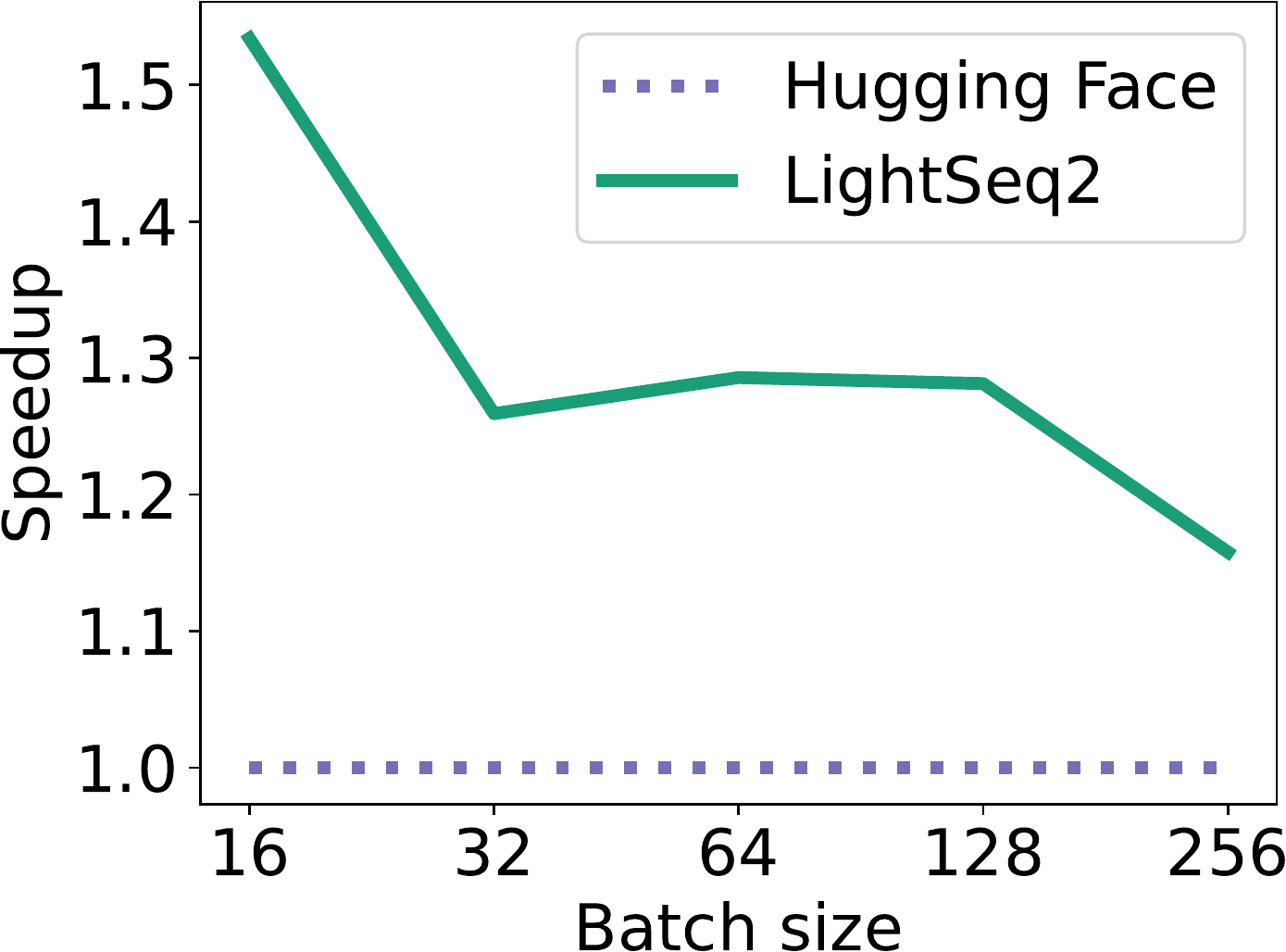}
}
\caption{\method speedup of Vision Transformer on image classification compared with Hugging Face.}
\label{fig:vit}
\end{figure}

\subsubsection{Transformer}\label{sec:eval:end2end:transformer}
We evaluate the performance of full Transformer on WMT14 English-German machine translation task. The baseline is Fairseq Transformer with Apex optimization. Fairseq implement Transformer with native Pytorch operators. Apex can further improve the performance of Fairseq Transformer by providing custom optimized operators.

We train transformer models at three different sizes (e.g., 6e6d represents six encoder and decoder layers) on eight V100/A100 GPUs. The batch token size denotes the total number of tokens in one mini-batch. Fig. \ref{fig:fp16_pt_8gpu_apex-ls} show that \method obtains a speedup of \textbf{1.4-2.8$\times$} on V100 and \textbf{1.5-3.5$\times$} on A100 compared to Fairseq Transformer with Apex optimization. 

Due to the computational graph optimizations and dependent reductions rewriting, \method achieves higher computational throughput and requires less GPU memory and I/O. As a result, \method obtains a higher speedup on A100 than V100. Moreover, under scenario of training large model, the bottleneck is memory bandwidth rather than computational throughput due to the limited batch size. \method also achieves higher speedup for larger model. To sum up, \method favors GPUs with higher peak throughput and larger models.

\begin{figure}[t]
\centering
\includegraphics[width=0.5\linewidth]{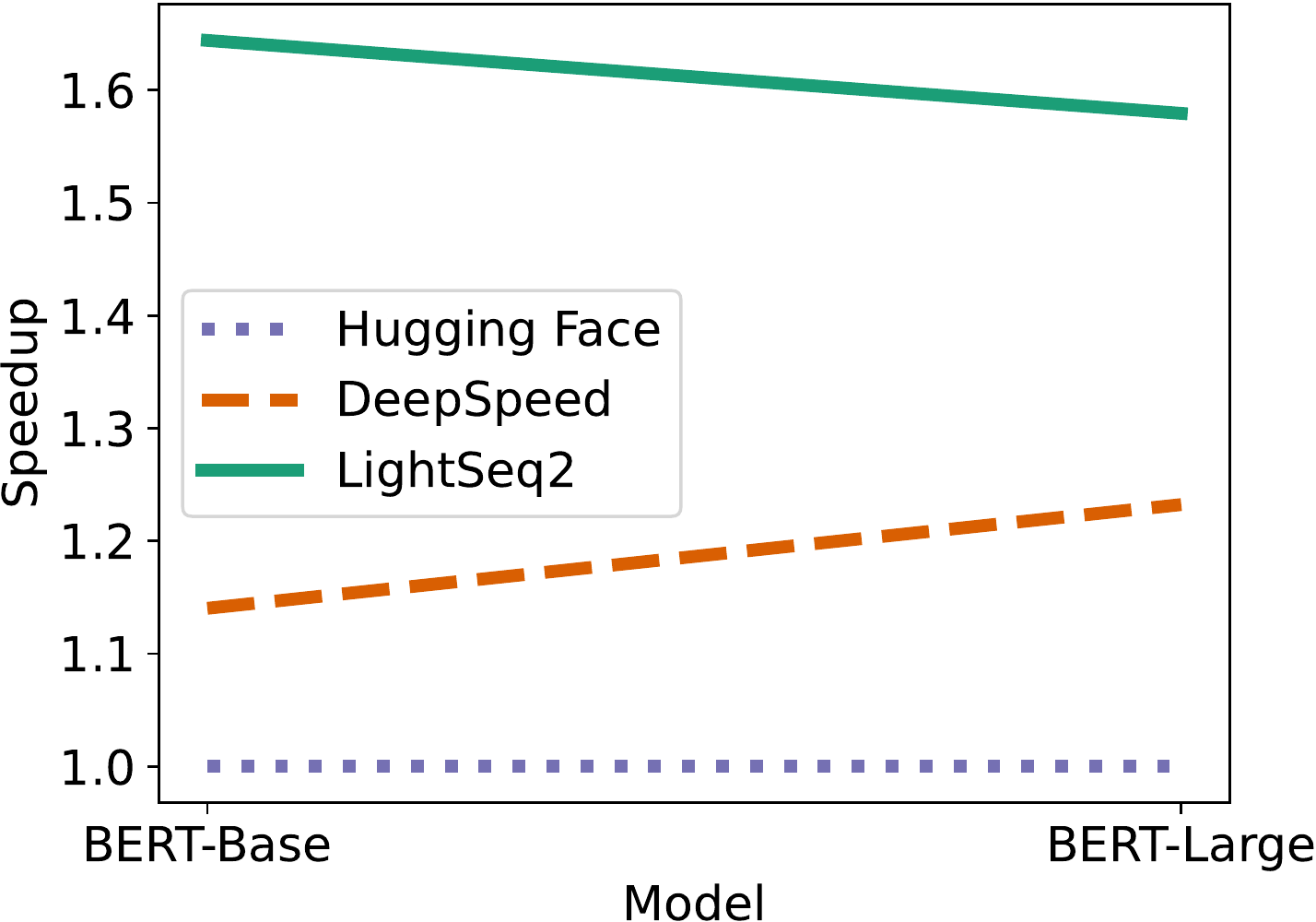}
\caption{Speedup on MRPC task using different BERT structures and eight V100 GPUs. Here we use samples per second to measure the speed.}
\label{fig:fp16-bert}
\end{figure}

\begin{figure}[t]
\centering
\subfigure[GPT-2 Base.] {
    \label{fig:gpt-2-base}
    \includegraphics[width=0.445\linewidth]{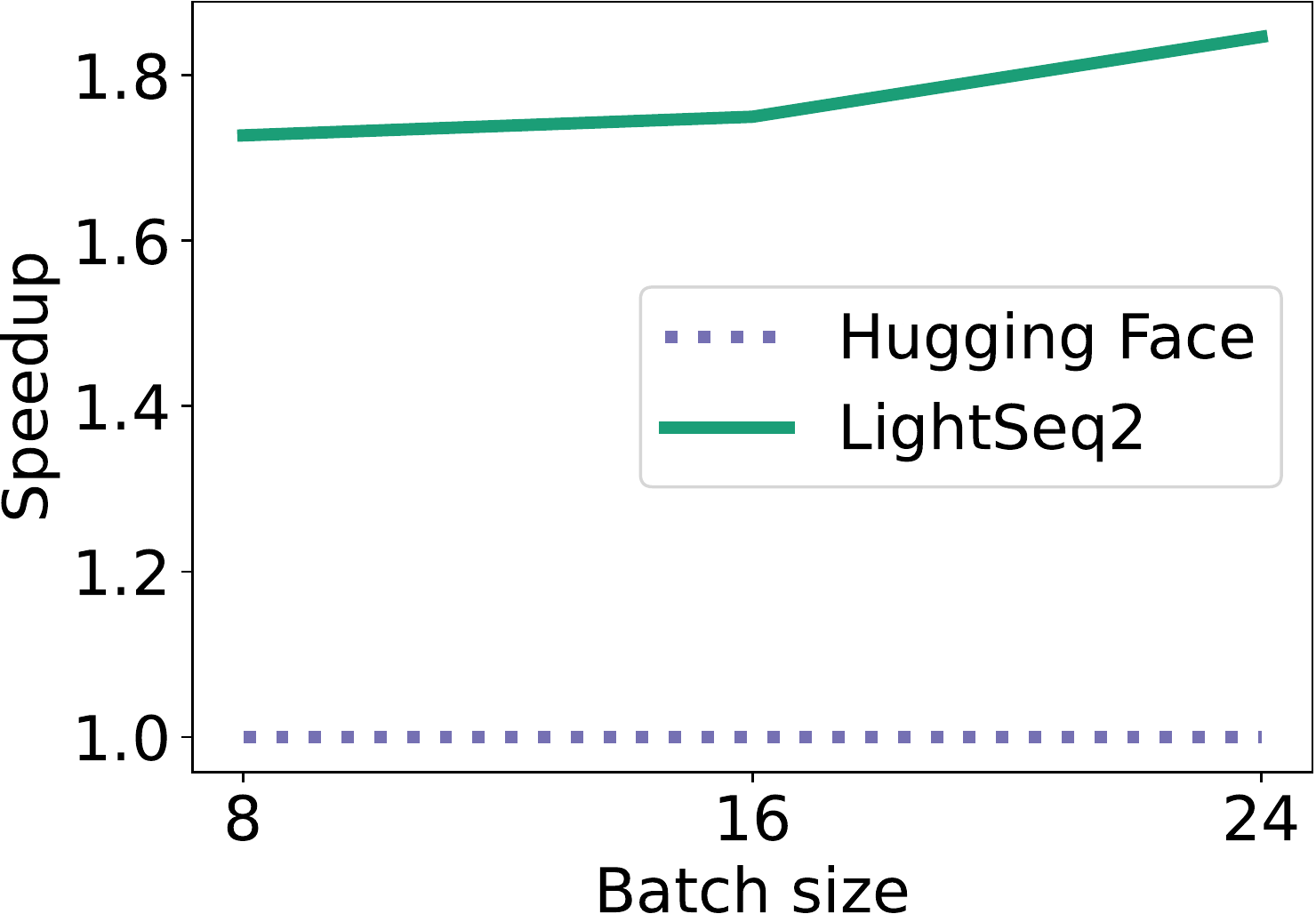}
}
\subfigure[GPT-2 Large.] {
    \label{fig:gpt-2-large}
    \includegraphics[width=0.445\linewidth]{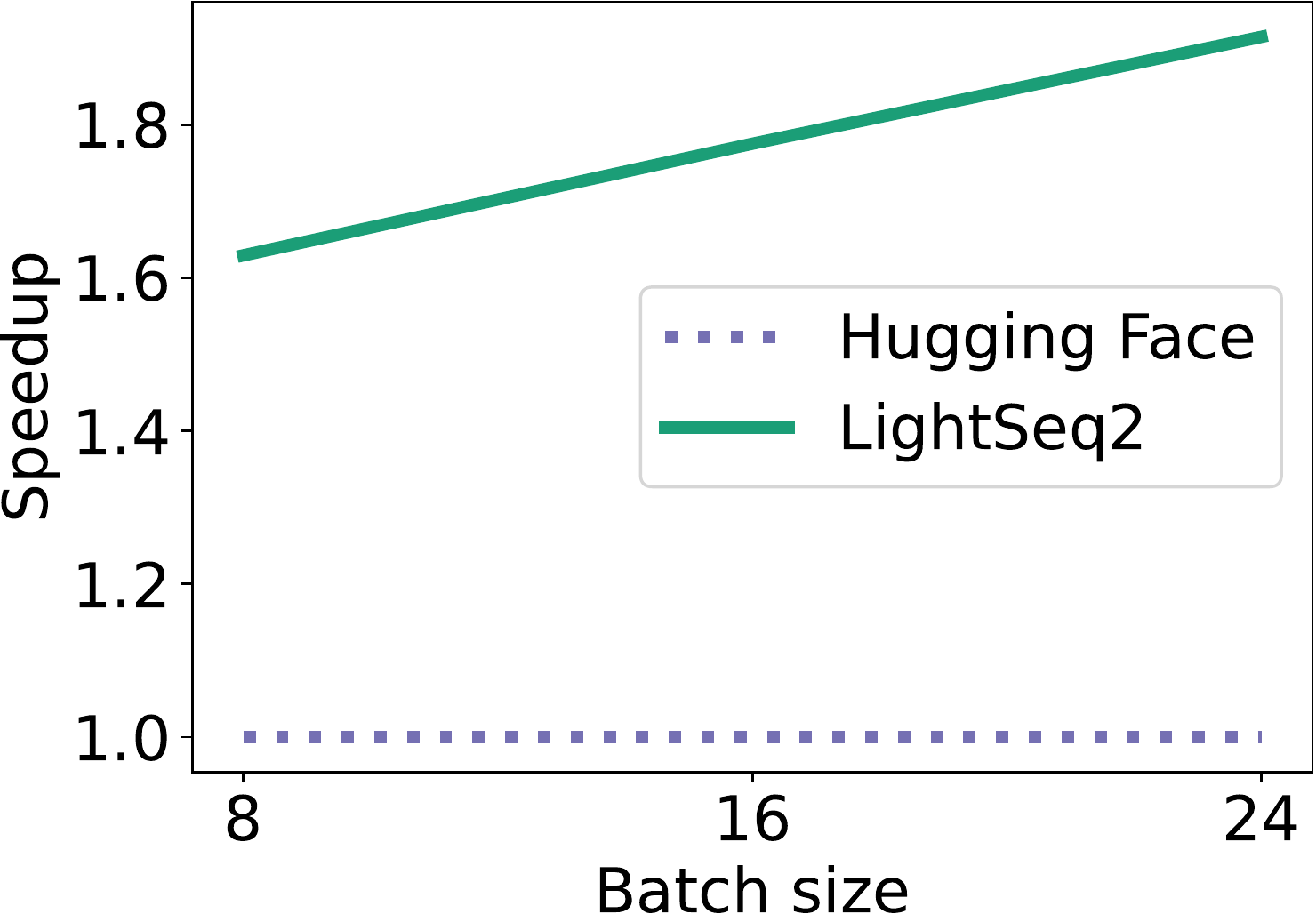}
}
\caption{\method speedup of GPT-2 compared with Hugging Face. GPT-2 Base model is trained on V100 and GPT-2 Large model is trained on A100.}
\label{fig:gpt-2}
\end{figure}

\begin{figure}[t]
\centering
\includegraphics[width=0.5\linewidth]{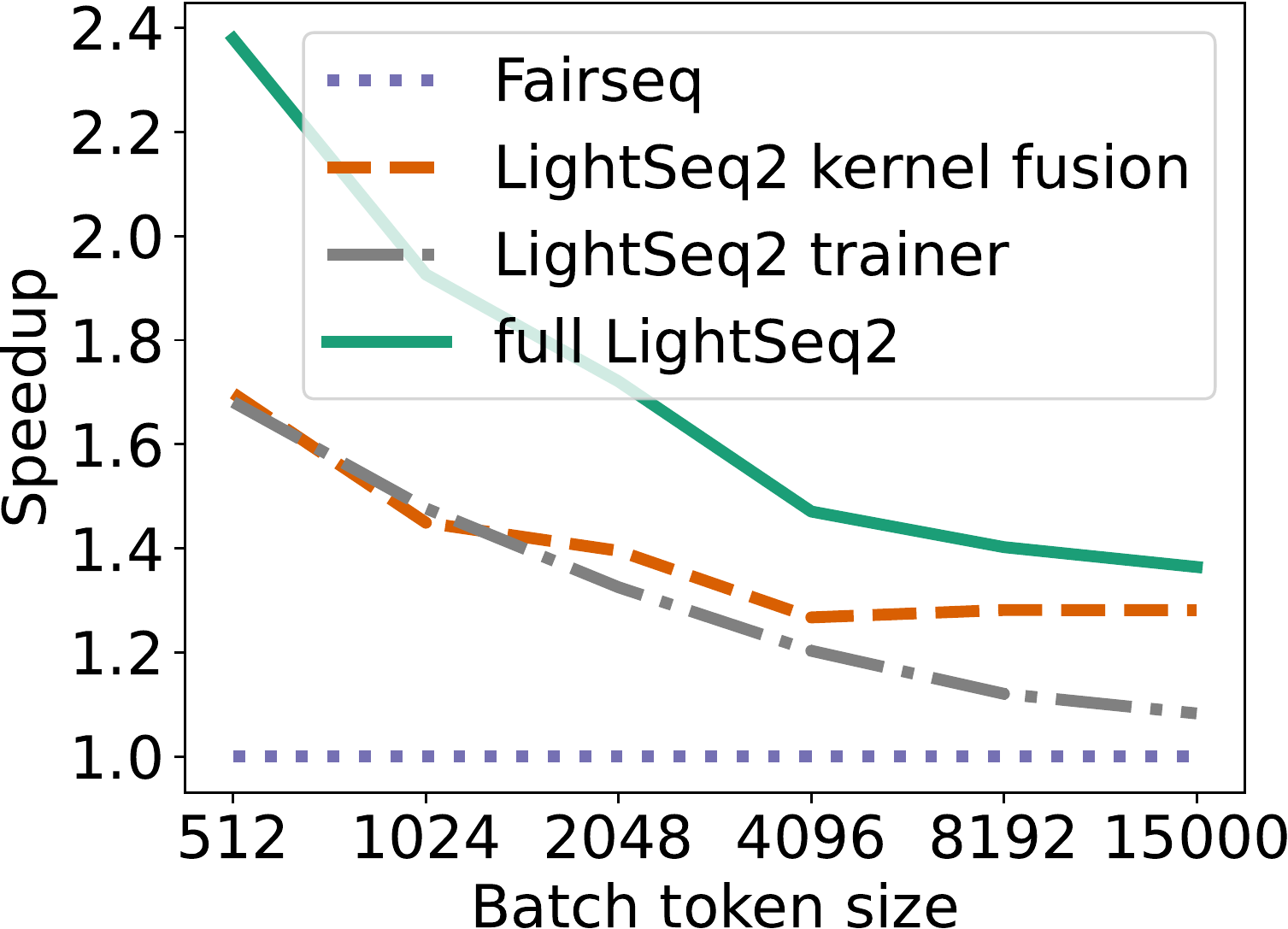}
\caption{\revise{Speedup for a 12-layer Transformer (6e6d) on 8 V100 GPUs. \method has three variations, with kernel fusion only, training only, and the full version.}}
\label{fig:kernel_cmp}
\end{figure}

\begin{figure*}[t]
\centering
\includegraphics[width=0.9\linewidth]{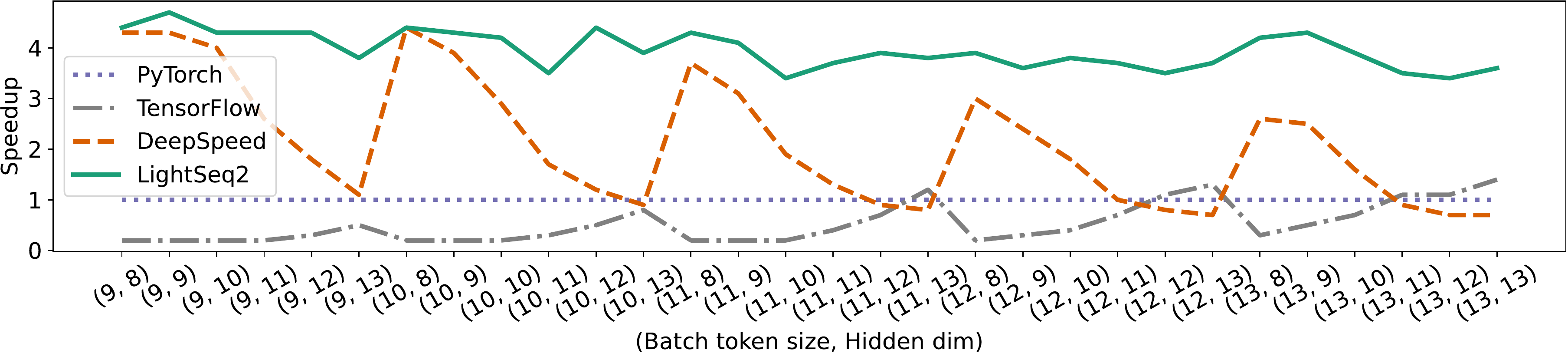}
\caption{Comparison between different implementations of \texttt{LayerNorm}. The $x$-axis coordinate values are all logarithms to 2 (e.g., $(12, 8)$ represents batch token size 4096 and hidden dimension 256).}
\label{fig:layernorm-fp16}
\end{figure*}

\begin{figure}[t]
\centering
\subfigure[Dropout.] {
    \label{fig:dropout-fp16}
    \includegraphics[width=0.445\linewidth]{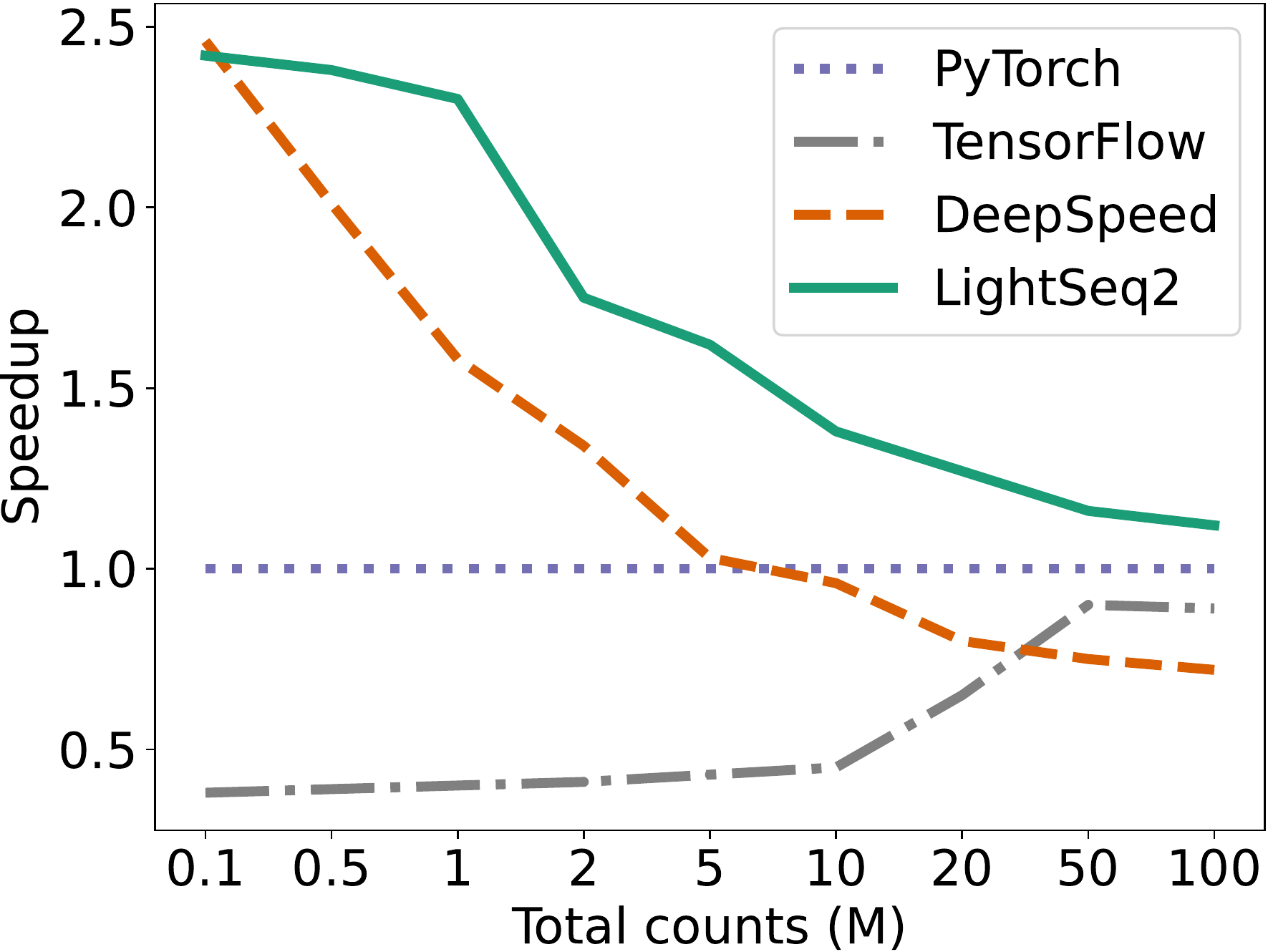}
}
\subfigure[Softmax.] {
    \label{fig:softmax-fp16}
    \includegraphics[width=0.421\linewidth]{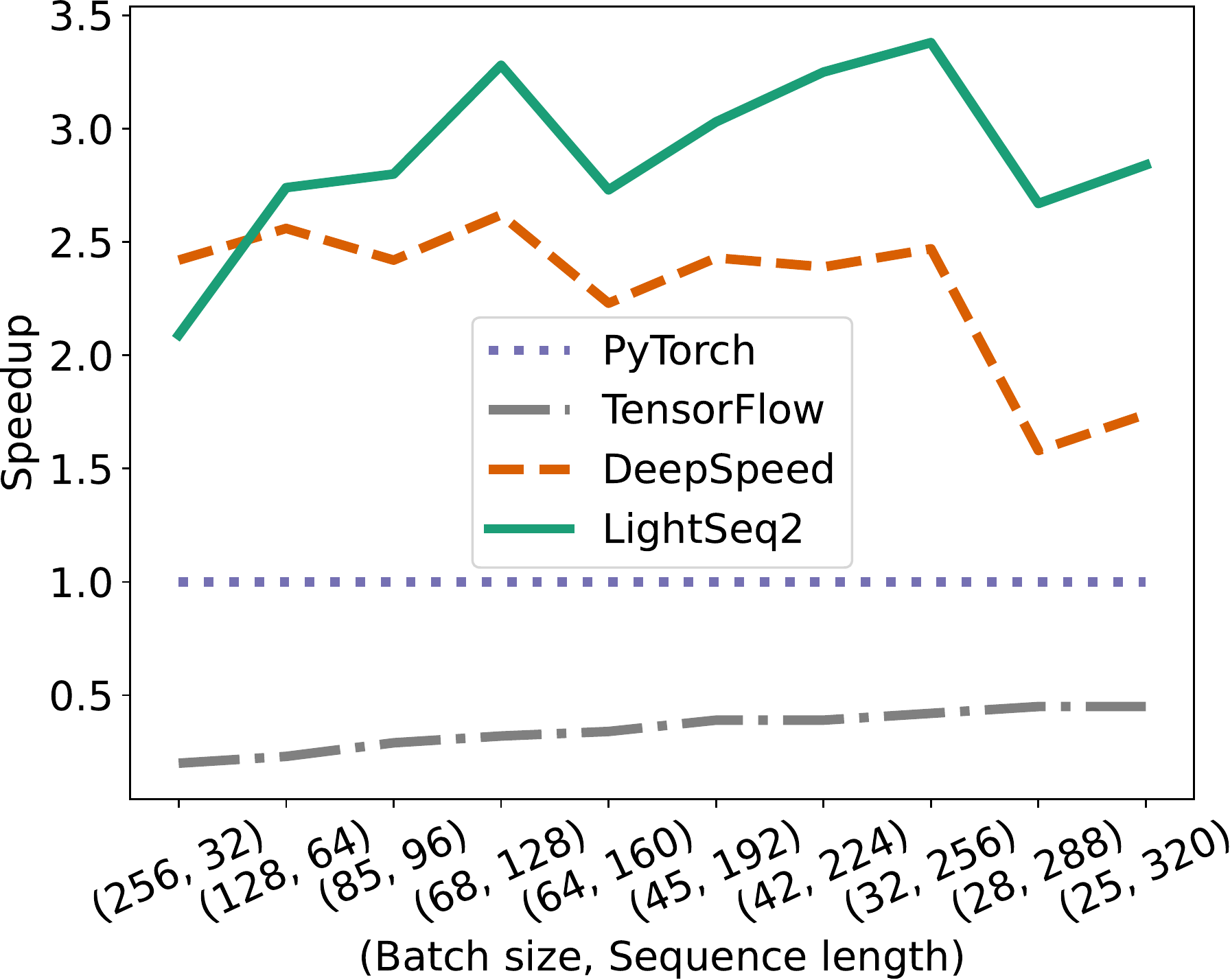}
}
\caption{Speedup for \texttt{Dropout} and \texttt{Softmax} operators.}
\label{fig:kernels}
\end{figure}

\subsubsection{ViT}
We evaluate the performance of vision Transformer (ViT)~\cite{dosovitskiy2021image} on image classification task using CIFAR-10 dataset. The baseline is Hugging Face ViT, which is based on native PyTorch operators. 

We train ViT models of two different sizes on eight V100 GPUs. ViT-B-32 and ViT-L-32 represent the base and large ViT models with patch size 32. The resolutions of the images are all $224 \times 224$. Thus the sequence lengths of the inputs of the ViT are all 50. \method can obtain a speedup of \textbf{1.2-1.7$\times$} on ViT-B-32 and \textbf{1.2-1.5$\times$} on ViT-L-32 compared to Hugging Face, as shown in Fig. \ref{fig:vit}.

The input shape of image classification task are quite different from machine translation. \method still outperforms the Hugging Face in all configurations by large margins, which verifies that our optimization is general for input shape and also suitable for computer vision tasks.

The ViT model only has an encoder with a relatively simple computational graph. With the increase of batch size, the proportion of time for matrix multiplication (GEMM) during training increases, which is not optimized by \method currently. As a result of it, the speedup decreases as batch size increases. When the batch size reaches the maximum (limited by a 32 GB of GPU memory), \method can still obtain a speedup of 1.2$\times$ on ViT-B-32 and ViT-L-32, which shows the efficiency of our optimization apart from GEMM.

\subsubsection{BERT}
We evaluate the performance of BERT (Transformer encoder only) on Microsoft Research Paraphrase Corpus (MRPC) task of General Language Understanding Evaluation (GLUE) benchmark. 

The baseline is Hugging Face BERT and DeepSpeed BERT. Hugging Face implements BERT with native Pytorch operators and is widely used due to its public pre-trained models. DeepSpeed\footnote{\url{https://github.com/microsoft/DeepSpeedExamples/tree/master/bing_bert}} optimizes the Transformer encoder layer and achieves the fastest performance for training BERT. For a fair comparison with DeepSpeed, we replace our efficient embedding, criterion, and trainer operators with theirs and individually verify the efficiency of our encoder layer.

We train BERT-Base and BERT-Large models on eight V100 GPUs and choose samples per second (SPS) as metric to evaluate performance. As shown in Fig. \ref{fig:fp16-bert}, \method obtains a speedup of \textbf{1.44$\times$} for BERT-Base and \textbf{1.28$\times$} for BERT-Large compared to DeepSpeed, which proves the efficiency of the \method encoder layer.

\subsubsection{GPT-2}
We evaluate the performance of GPT-2 (Transformer decoder only) on language modeling task using WikiText \cite{wikitext} dataset. Our baseline is Hugging Face GPT-2, which is based on native PyTorch operators.

We train GPT-2 Base with 117M parameters on eight V100 GPUs and GPT-2 Large with 762M parameters on eight A100 GPUs. Iterations per second is chosen as the metric to evaluate performance. As shown in Fig. \ref{fig:gpt-2}, \method obtains a speedup of \textbf{1.7-1.8$\times$} for GPT-2 Base on V100 and \textbf{1.6-1.9$\times$} for GPT-2 Large on A100 compared to Hugging Face, which proves the efficiency of the \method decoder layer.

\subsection{Speedup Breakdown}\label{sec:eval:ablation}
To demonstrate the benefits of our optimizations in detail, we test the performance of \method from operator and layer level on one NVIDIA Tesla V100 GPU. \revise{To measure the speed, we run each operator/layer for 10 times and take the average time.}

\subsubsection{Operator-wise Speedup}
We choose four common operations, \texttt{Dropout} (element-wise), \texttt{Adam} (element-wise), \texttt{Softmax} (reduction) and \texttt{LayerNorm} (reduction), to show the efficient of CUDA kernels from \method. Our baselines are their implementations from PyTorch, TensorFlow and DeepSpeed.

Fig. \ref{fig:layernorm-fp16} shows the performance of \method \texttt{LayerNorm}. \method gains a speedup of about \textbf{4$\times$} despite of the batch token size and hidden dimension. However, with the increase of batch token size or hidden dimension, the speedup of DeepSpeed drops significantly. Our technique shows high performance even when the original kernel has limited parallelism. If the number of elements is huge, the speed of DeepSpeed is not even as good as PyTorch. On the other hand, TensorFlow is not as fast as PyTorch in most cases, except when there are too many elements.

\begin{figure}[t]
\centering
\subfigure[Adam.] {
    \label{fig:v100-adam}
    \includegraphics[width=0.445\linewidth]{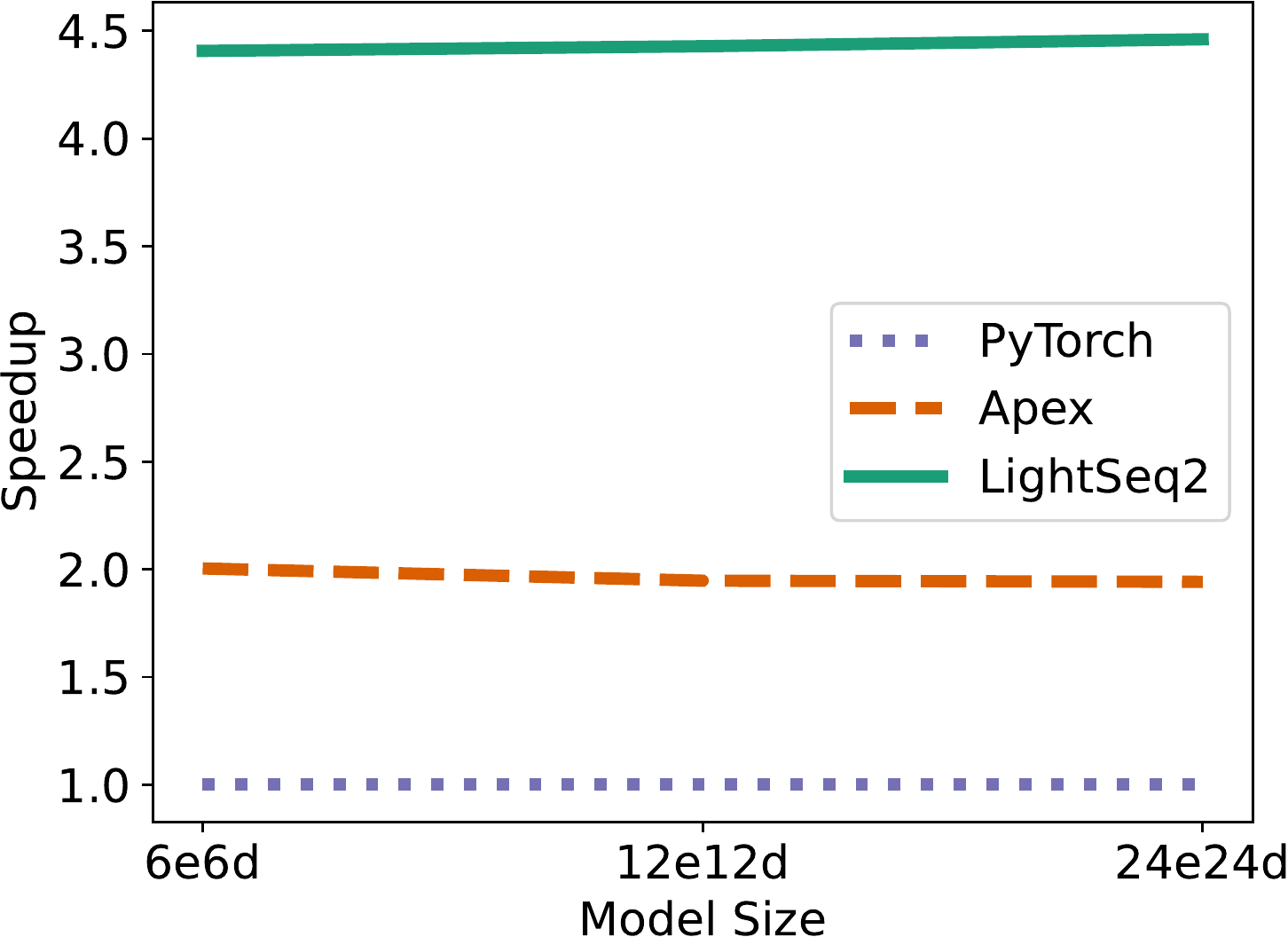}
}
\subfigure[\revise{SGD.}] {
    \label{fig:v100-sgd}
    \includegraphics[width=0.445\linewidth]{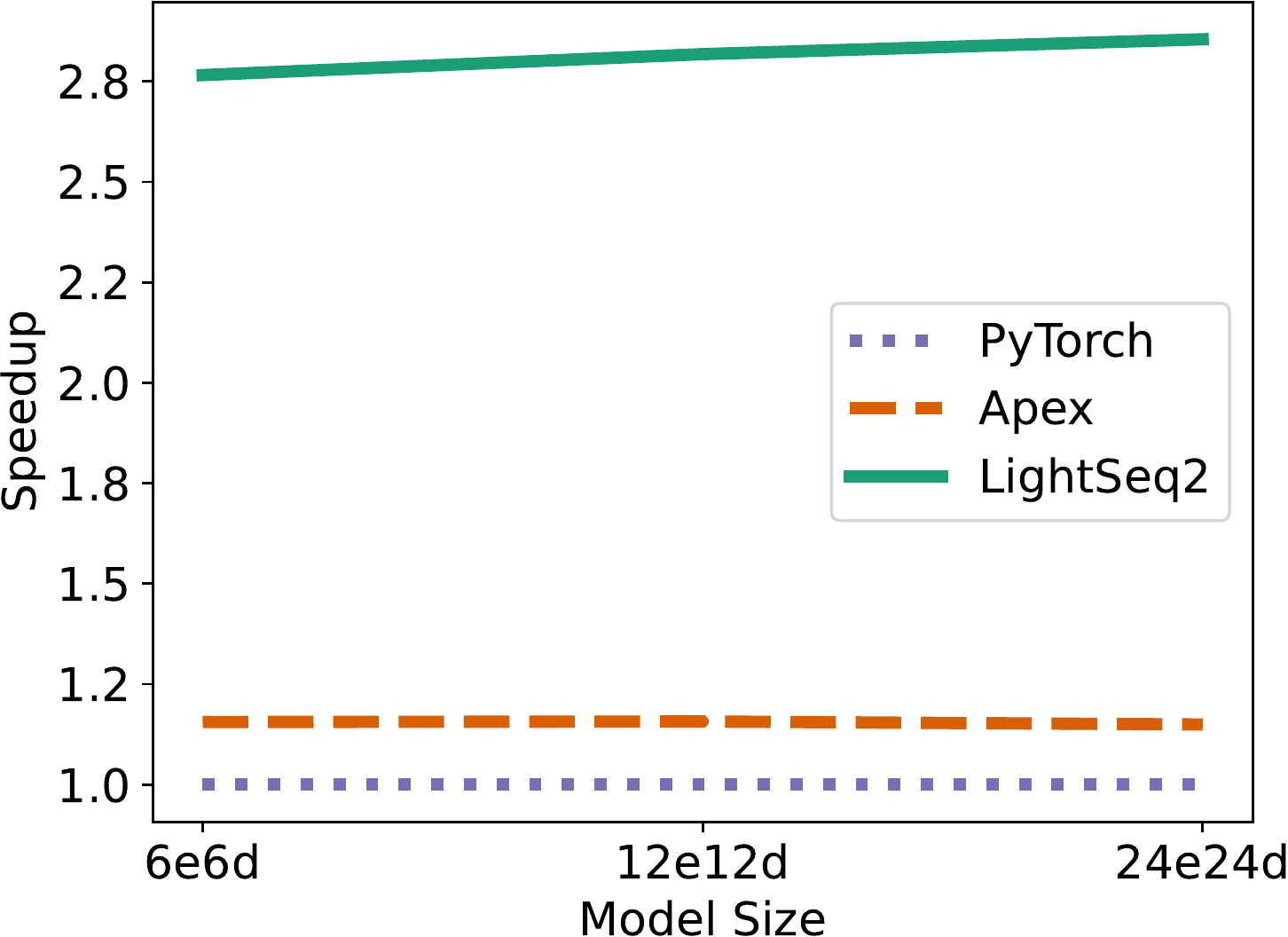}
}
\caption{\revise{Speedup of trainer. Notice that \method gains a consistent speedup of \textbf{2.3$\times$} for \texttt{Adam} and \textbf{2.4$\times$} for \texttt{SGD} over Apex across different model sizes.}}
\label{fig:v100-trainer}
\end{figure}

\begin{figure}[t]
\centering
\subfigure[Forward.] {
    \label{fig:layer-fw}
    \includegraphics[width=0.46\linewidth]{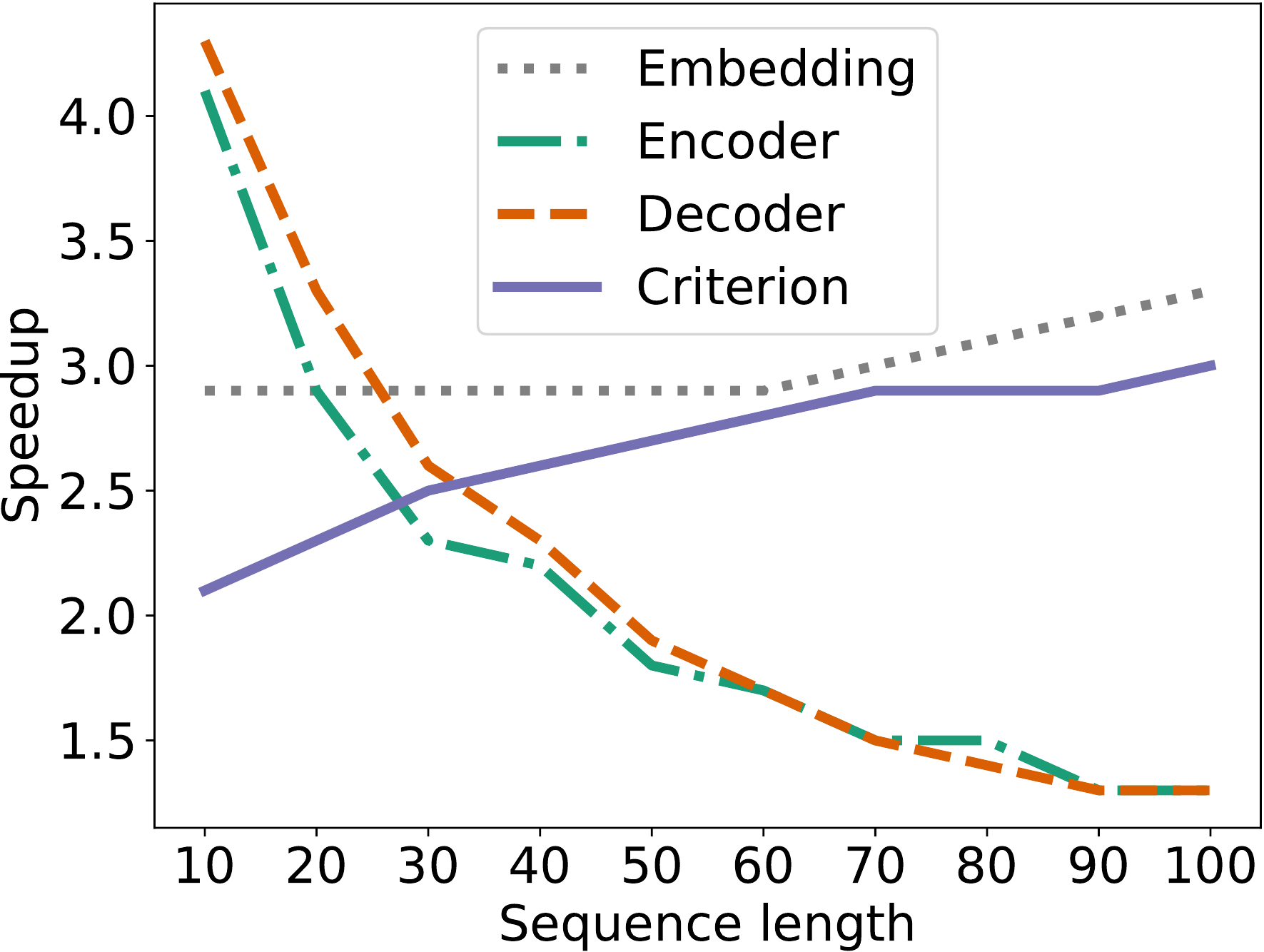}
}
\subfigure[Backward.] {
    \label{fig:layer-bw}
    \includegraphics[width=0.46\linewidth]{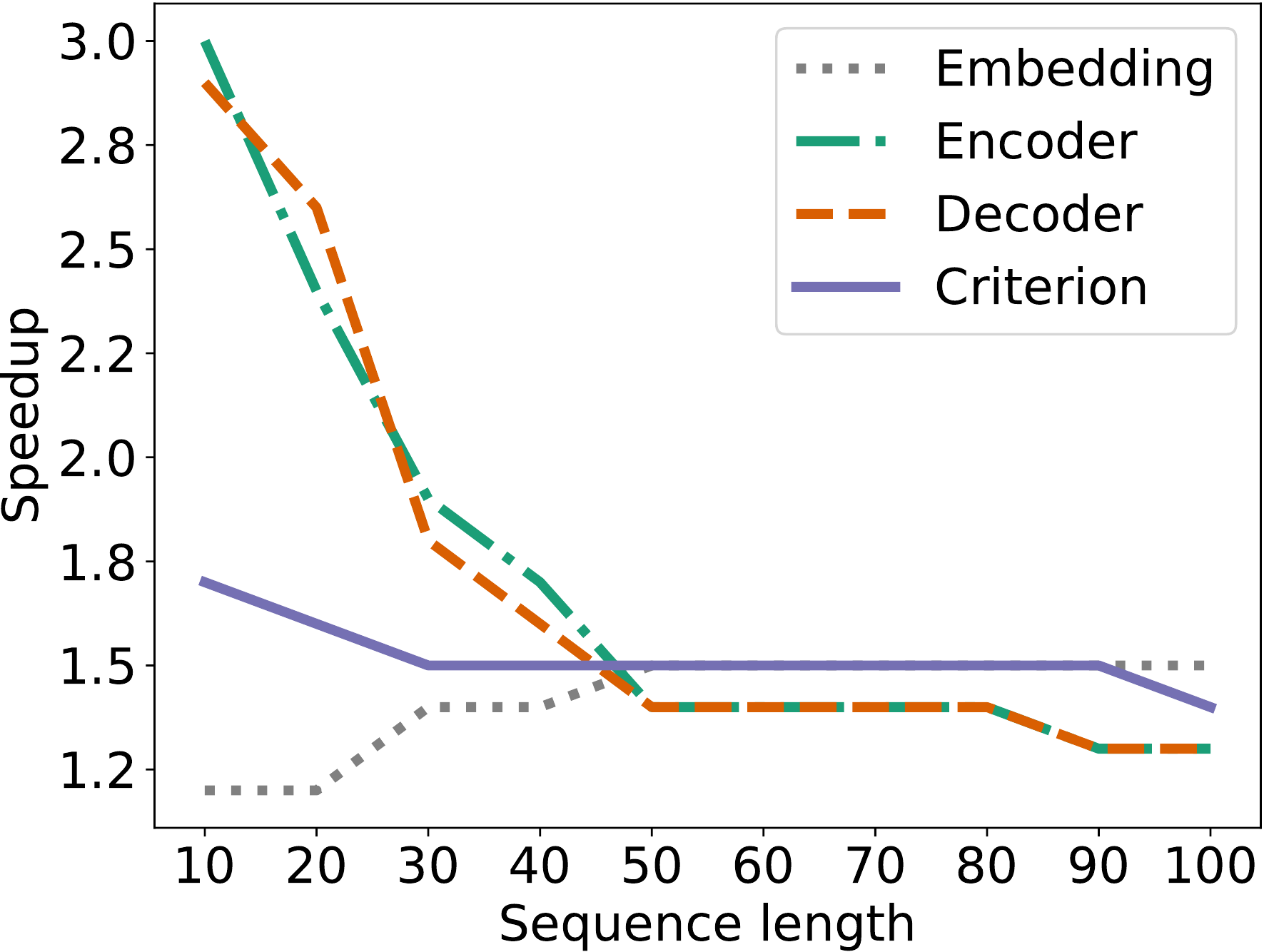}
}
\caption{\method speedup of different layers.}
\label{fig:layer-speedup}
\end{figure}

\begin{figure}[t]
\centering
\subfigure[Transformer-Base.] {
    \label{fig:gpu_mem_base}
    \includegraphics[width=0.42\linewidth]{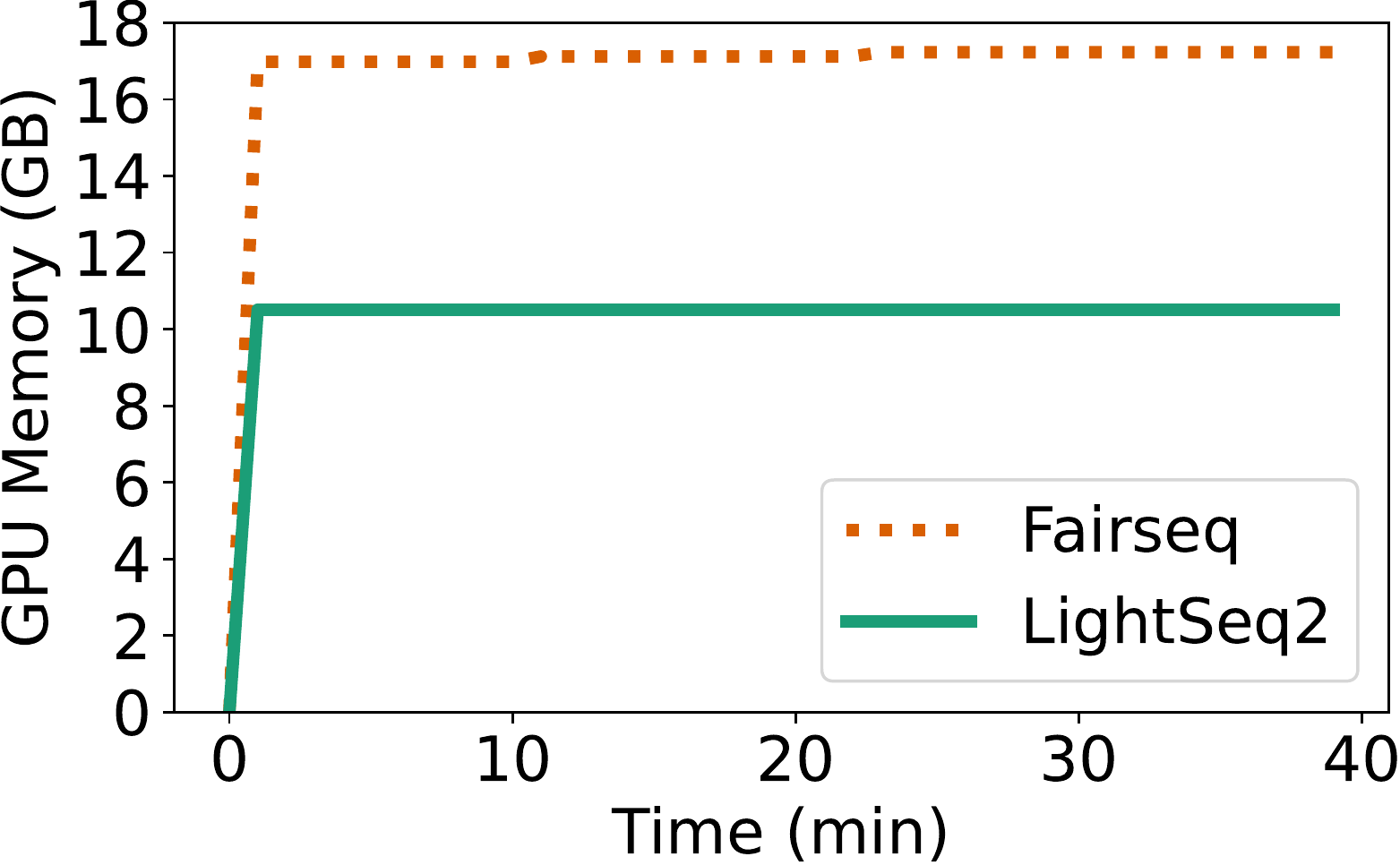}
}
\subfigure[Transformer-Big.] {
    \label{fig:gpu_mem_big}
    \includegraphics[width=0.42\linewidth]{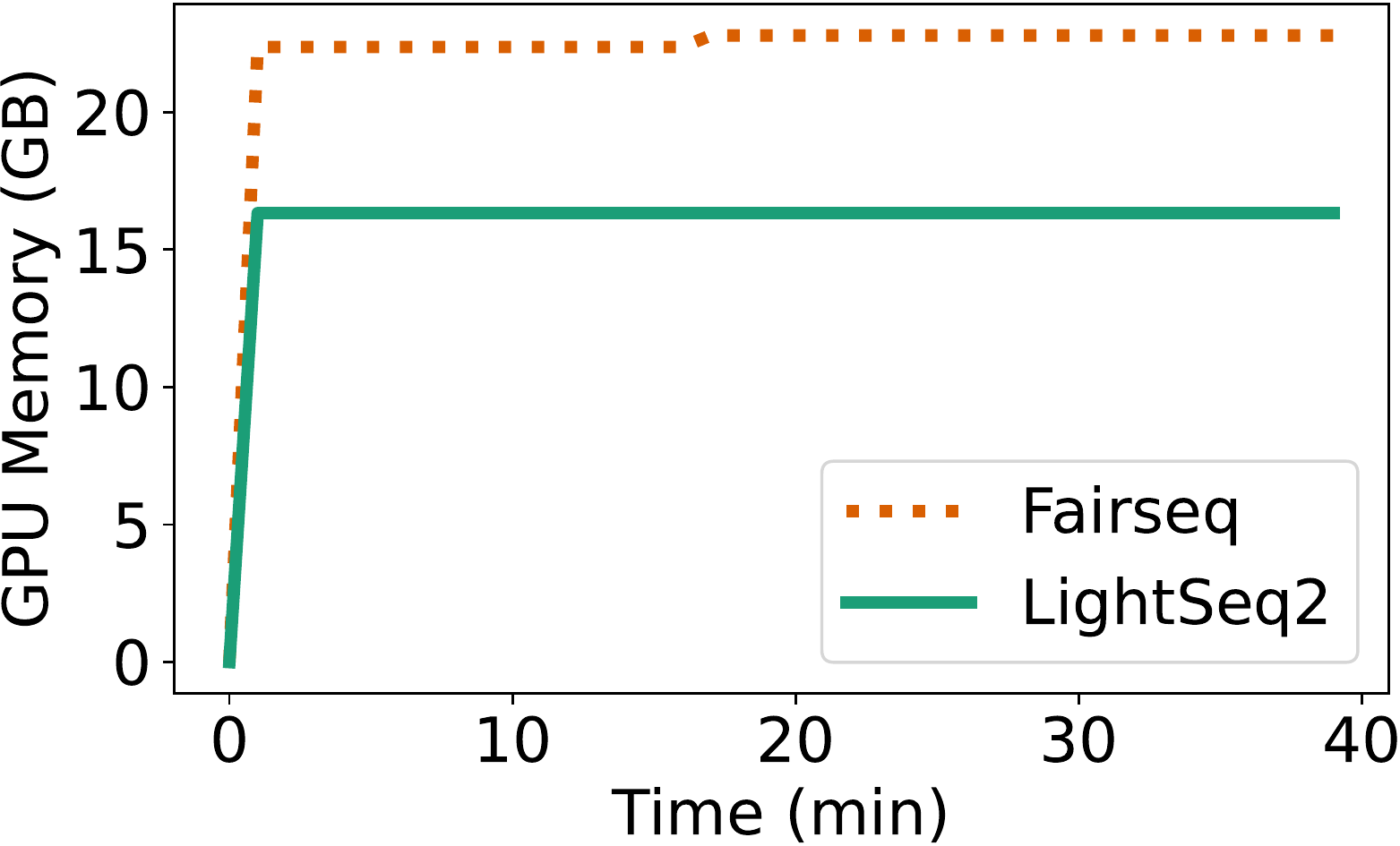}
}
\caption{GPU memory usage for machine translation on one V100.}
\label{fig:gpu_mem}
\end{figure}

\begin{figure}[t]
\centering
\subfigure[Transformer-Base.] {
    \label{fig:gpu_percent_base}
    \includegraphics[width=0.445\linewidth]{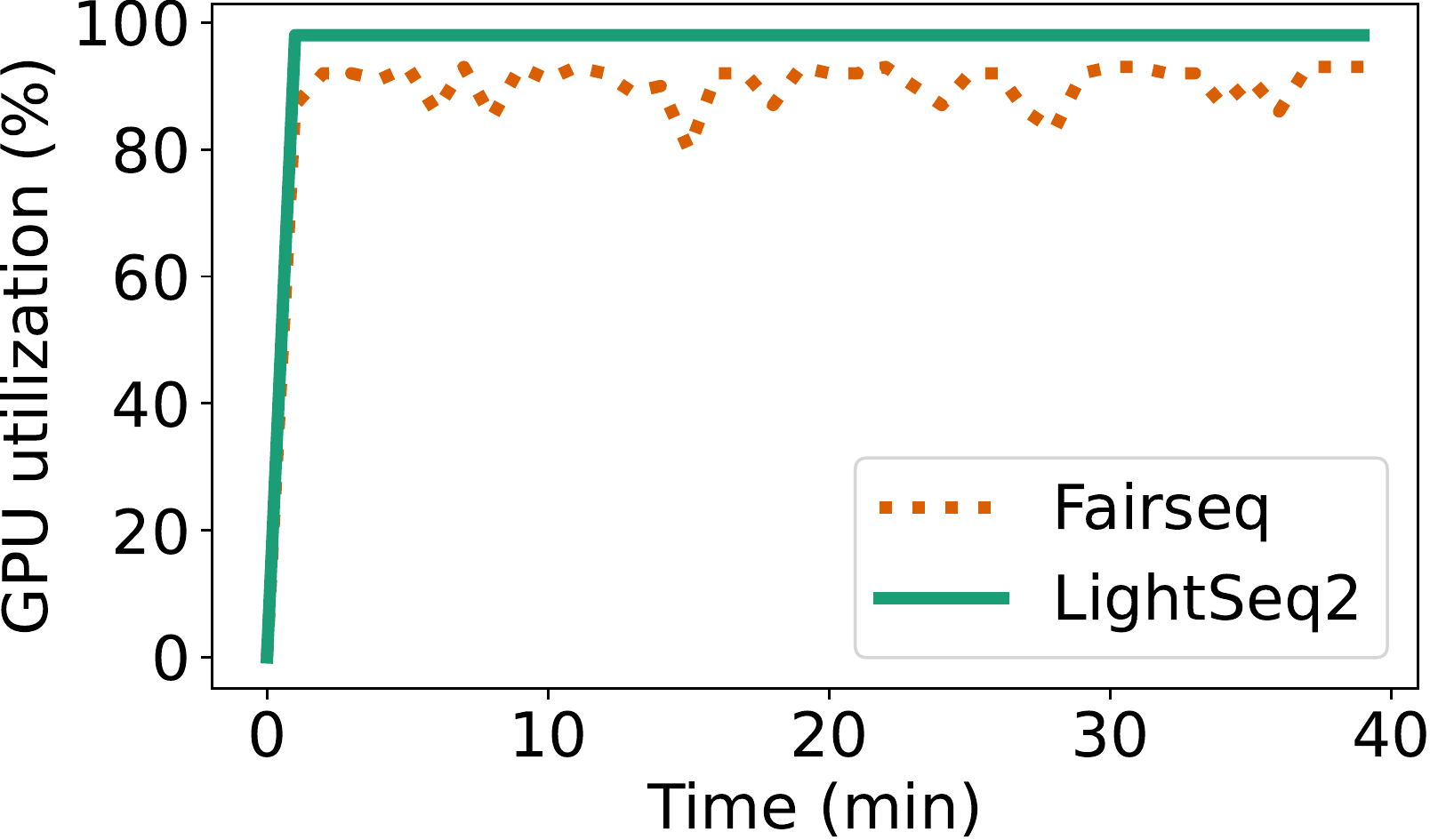}
}
\subfigure[Transformer-Big.] {
    \label{fig:gpu_percent_big}
    \includegraphics[width=0.445\linewidth]{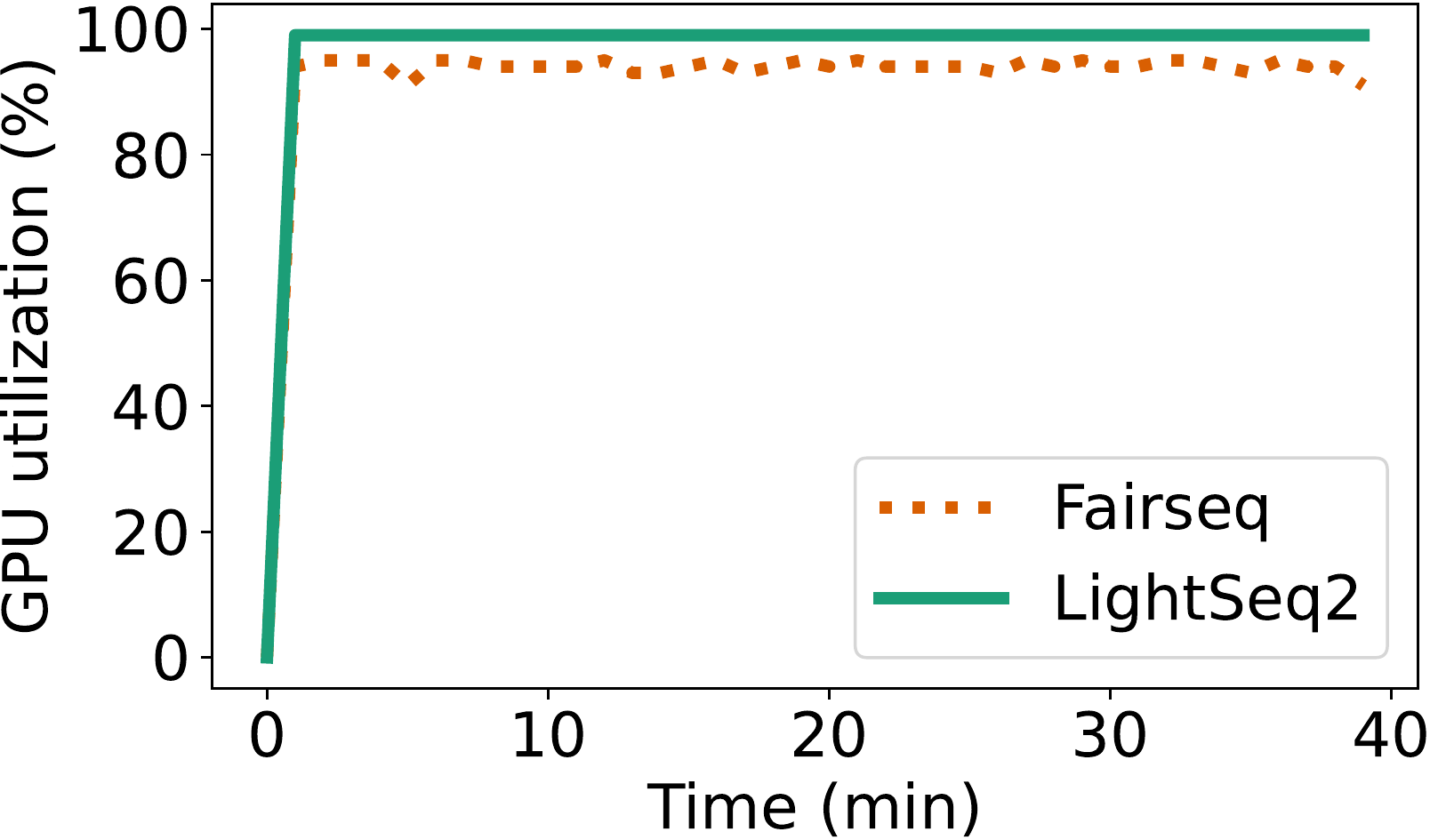}
}
\caption{GPU utilization for machine translation on one V100.}
\label{fig:gpu_percent}
\end{figure}

Fig. \ref{fig:dropout-fp16} shows the performance of \method \texttt{Dropout}. As the number of elements increases, both DeepSpeed and \method become slower. When the number of elements is greater than five million, DeepSpeed becomes slower than PyTorch. The gap between TensorFlow and PyTorch becomes smaller, but \method still has a speedup of \textbf{1.2-1.5$\times$}.

Fig. \ref{fig:softmax-fp16} shows the performance of \method \texttt{Softmax}. Unlike the other kernels, as the batch size and sequence length increase, the speedup of \method becomes larger, due to the specific optimization for different input shapes. The trends of DeepSpeed and TensorFlow are similar to the other kernels.

Fig. \ref{fig:v100-trainer} shows the performance of \method trainer. \method gains a consistent speedup of \textbf{2.3$\times$} for \texttt{Adam} and \textbf{2.4$\times$} for \texttt{SGD} over Apex despite of model size, which demonstrates the benefits of the symbolic tensor link and on-the-fly conversion as discussed in Section~\ref{sec:approch-adam}. 

\revise{
Finally, as shown in Fig. \ref{fig:kernel_cmp}, we evaluate the speed of only using kernel fusion or \method trainer. As the batch token size increases, the speedup decreases due to the growing proportion of GEMM kernels. The gap between kernel fusion and \method trainer becomes larger as the batch token size increases, which is consistent with Fig. \ref{fig:time_dist}. This is due to the fact that the forward and backward processes dominate as the batch token size grows.
}

\subsubsection{Layer-wise Speedup}
We evaluate the speedup of \method layers over PyTorch implementations (Fairseq). The layer configurations are the same as in Transformer-Big models.

The results of forward and backward propagation are in Fig. \ref{fig:layer-speedup}. We can draw the following conclusions:
\begin{itemize}
    \item \method obtains higher speedup in forward propagation than in backward. This is because the time of backward propagation contains the part of gradient copies.
    \item The speedup ratios of the encoder and the decoder decrease rapidly as the sequence length becomes larger. However, the speedups of embedding and criterion are stable. This is because when the sequence length increases, the large matrix multiplication in the encoder and the decoder quickly saturates the GPU parallelism, leading to a sub-linear improvement.
\end{itemize}

In all cases, \method layers are faster than PyTorch, which demonstrates the benefits of computational graph optimizations as discussed in Section~\ref{sec:approch-graph-optimize}.

\subsection{Benefit of Memory Optimization}\label{sec:eval:mem-opt}


We evaluate the memory efficiency on WMT14 English-German machine translation task using V100 GPU. We use Fairseq as baseline, both using a batch token size of 8192. All experiments ran for 40 minutes to measure under a stable state.

Fig. \ref{fig:gpu_mem} illustrates the GPU memory occupancy of both Transformer-Base (6e6d, 512d, 8 heads) and Transformer-Big (6e6d, 1024d, 16 heads) models. Fairseq consumes about 6 GB more GPU memory than \method in both cases. For example, Transformer-Base models based on Fairseq can not run on a GPU with only 16 GB memory. Another phenomenon is that the GPU memory of Fairseq will gradually increase as it runs. This is because Fairseq dynamically allocates and releases the GPU memory when the sequence lengths differ. Thus, Fairseq needs to apply for additional GPU memory if a long sequence is input into the model. In contrast, \method allocates the maximal GPU memory in advance, so there will be no memory change during the training. As shown in Fig. \ref{fig:fp16_pt_8gpu_apex-ls}, \method can train deep models under large batch size (batch size 15k for 6e6d Transformer and batch size 8192 for 12e12d Transformer on 32 GB V100), whereas Fairseq Transformer encounters out-of-memory error, which is useful for training deep models.

Fig. \ref{fig:gpu_percent} illustrates the GPU utilization of both Transformer-Base and Transformer-Big models. In the whole training process, \method keeps a utilization rate of about 99\% in both cases. However, for Fairseq, the utilization of the Transformer-Base model is very unstable. The lowest is only 80\%, and in most cases, it fluctuates between 87\% and 93\%, mainly due to frequent memory allocation and release. The utilization of the Transformer-Big model is much more stable, but the highest is only 95\%.

If the batch token size is smaller, the gap between the two implementations will be more obvious. For example, when the batch token size is reduced to 4096, the memory utilization of the Transformer-Base model based on Fairseq is only 73\%, while the \method is as high as 96\%.

\begin{figure}[t]
\centering
\subfigure[] {
    \label{fig:a100-gpuscale}
    \includegraphics[width=0.42\linewidth]{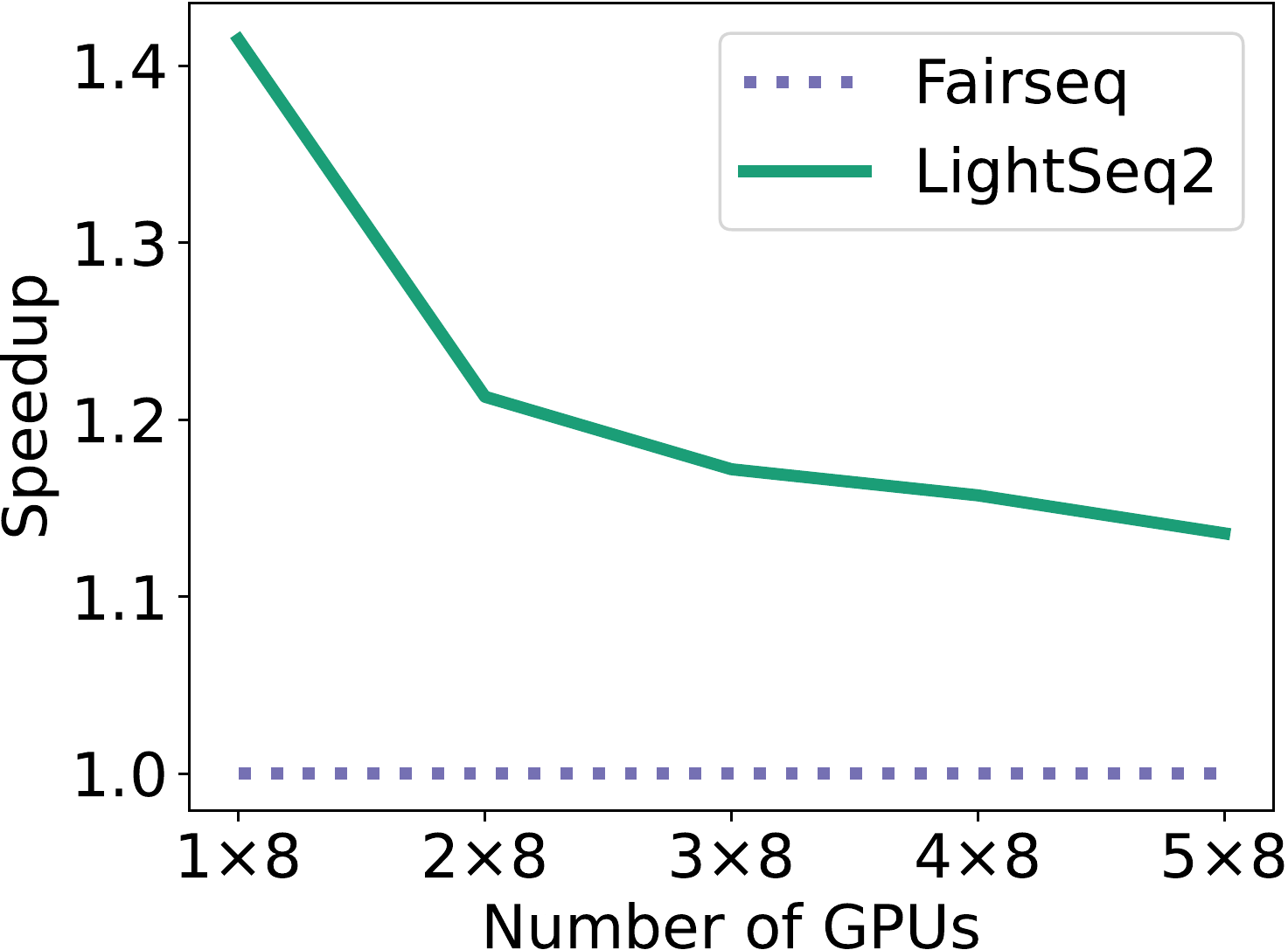}
}
\subfigure[] {
    \label{fig:a100-modelscale}
    \includegraphics[width=0.45\linewidth]{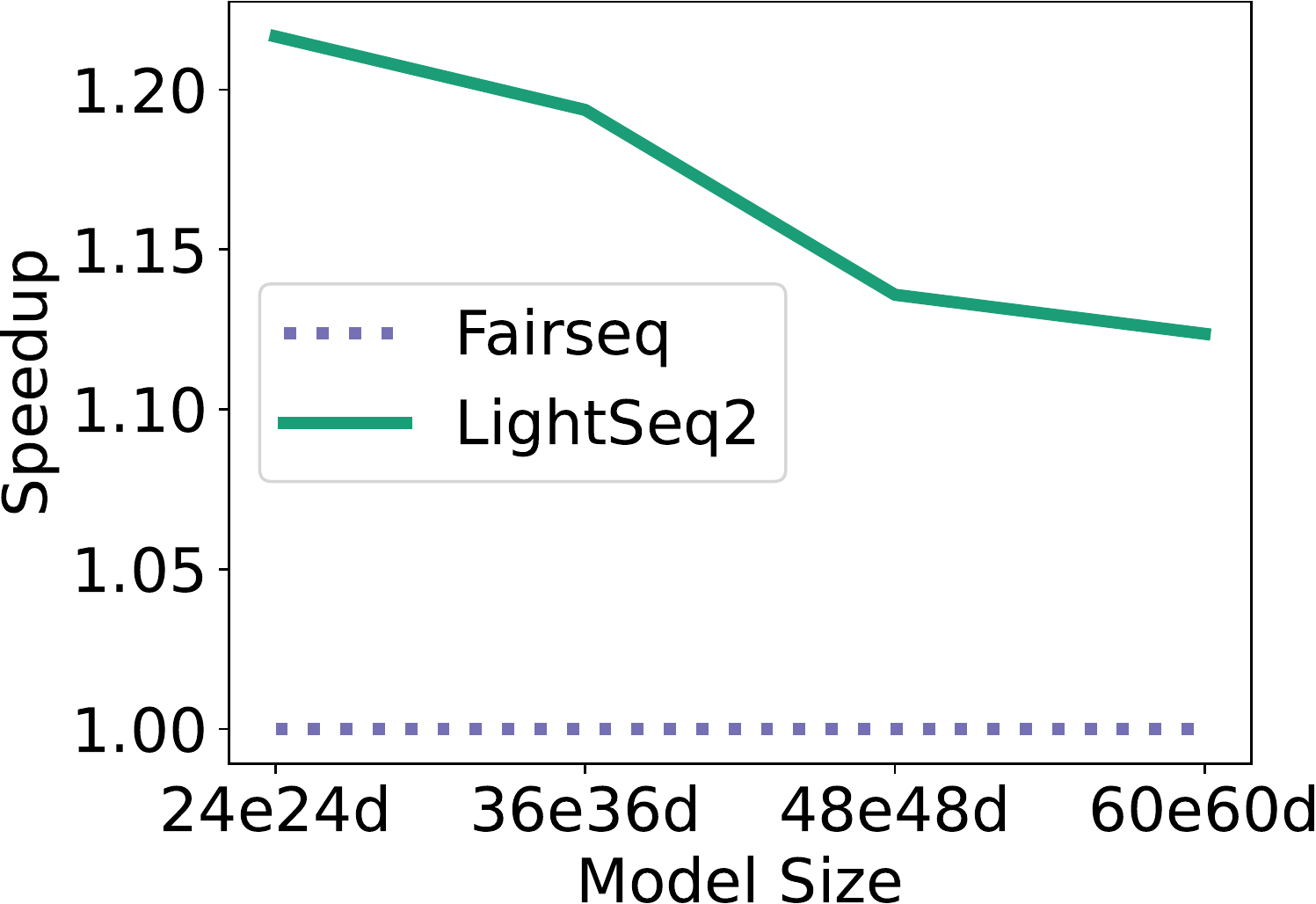}
}
\caption{\revise{Scalability of \method. (a) Speedup on different numbers of A100 GPUs for 48e48d model. Here $5 \times 8$ represents five nodes with eight GPUs on each (40 GPUs in total). (b) Speedup for varying sizes of models on $5 \times 8$ A100 GPUs. 60e60d has a total of 120 Transformer layers.}  }
\label{fig:multigpu}
\end{figure}

\revise{\subsection{Scalability}\label{sec:eval:scalability}}
\revise{
To evaluate the scalability of \method,
we extend the experiments in Section~\ref{sec:eval:end2end:transformer} to multiple nodes, each with 8 A100 GPUs. 
\method provides specific optimizations for the computation of Transformer and general optimizations for popular trainers such as \texttt{Adam} and \texttt{SGD}. 
We adopt the same all-reduce synchronization strategy from PyTorch. 
As the number of nodes increases, \method obtains a speedup of \textbf{1.14-1.41$\times$} for 48e48d model on different numbers of GPUs (Fig. \ref{fig:a100-gpuscale}) and \textbf{1.12-1.22$\times$} on $5 \times 8$ GPUs for different sizes of models (Fig. \ref{fig:a100-modelscale}).
These results show that \method scale well and gain consistent speedup for distributed training scenario and for increasing model sizes. 

With the increase of number of GPUs or model sizes, the proportion of time for synchronization during training increases. 
As a result, the speedup decreases as number of GPUs or model size increases. 
Nevertheless, \method still obtains a speedup of 1.12$\times$ on $5 \times 8$ A100 GPUs for a large Transformer model with a total of 120 layers (60e60d), which shows the system's capability for large model on many GPUs. 
In the future, \method can be further integrated with synchronization optimization techniques such as gradient compression, which are orthogonal to the focus of the current paper. 
}

\section{Conclusion}
\label{sec:conclusion}
In this paper, we describe a series of engineering-based GPU optimization techniques for fast Transformer training. Compared with existing approaches, our system strictly follows the standard training algorithm, therefore guarantees the quality and reproducibility of existing models. We systematically compare our work with existing state-of-the-art systems with various settings and analyze each component's performance, demonstrating the solidness and scalability of the contribution. Compared to PyTorch and TensorFlow implementations, \method can obtain a speedup of up to 3$\times$ under different configurations. 

In the future, there are still room for additional optimization such as padding removing and better memory management strategy. 


\section*{Acknowledgment}
We would like to thank the colleagues in Volctrans team for their contributions in project development. We also thank the teachers and classmates at University of California Santa Barbara for their suggestions on paper writing.

\bibliographystyle{IEEEtran}
\bibliography{paperref}

\begin{thebibliography}{10}
\providecommand{\url}[1]{#1}
\csname url@samestyle\endcsname
\providecommand{\newblock}{\relax}
\providecommand{\bibinfo}[2]{#2}
\providecommand{\BIBentrySTDinterwordspacing}{\spaceskip=0pt\relax}
\providecommand{\BIBentryALTinterwordstretchfactor}{4}
\providecommand{\BIBentryALTinterwordspacing}{\spaceskip=\fontdimen2\font plus
\BIBentryALTinterwordstretchfactor\fontdimen3\font minus
  \fontdimen4\font\relax}
\providecommand{\BIBforeignlanguage}[2]{{%
\expandafter\ifx\csname l@#1\endcsname\relax
\typeout{** WARNING: IEEEtran.bst: No hyphenation pattern has been}%
\typeout{** loaded for the language `#1'. Using the pattern for}%
\typeout{** the default language instead.}%
\else
\language=\csname l@#1\endcsname
\fi
#2}}
\providecommand{\BIBdecl}{\relax}
\BIBdecl

\bibitem{NIPS2017_3f5ee243}
A.~Vaswani, N.~Shazeer, N.~Parmar, J.~Uszkoreit, L.~Jones, A.~N. Gomez,
  L.~Kaiser, and I.~Polosukhin, ``Attention is all you need,'' in \emph{Proc.
  of NeurIPS}, I.~Guyon, U.~von Luxburg, S.~Bengio, H.~M. Wallach, R.~Fergus,
  S.~V.~N. Vishwanathan, and R.~Garnett, Eds., 2017, pp. 5998--6008.

\bibitem{DBLP:conf/naacl/DevlinCLT19}
J.~Devlin, M.-W. Chang, K.~Lee, and K.~Toutanova, ``{BERT}: Pre-training of
  deep bidirectional transformers for language understanding,'' in \emph{Proc.
  of NAACL-HLT}, 2019, pp. 4171--4186.

\bibitem{gpt2}
A.~Radford, J.~Wu, R.~Child, D.~Luan, D.~Amodei, and I.~Sutskever, ``Language
  models are unsupervised multitask learners,'' 2018.

\bibitem{DBLP:conf/nips/YangDYCSL19}
Z.~Yang, Z.~Dai, Y.~Yang, J.~G. Carbonell, R.~Salakhutdinov, and Q.~V. Le,
  ``Xlnet: Generalized autoregressive pretraining for language understanding,''
  in \emph{Proc. of NeurIPS}, H.~M. Wallach, H.~Larochelle, A.~Beygelzimer,
  F.~d'Alch{\'{e}}{-}Buc, E.~B. Fox, and R.~Garnett, Eds., 2019, pp.
  5754--5764.

\bibitem{DBLP:journals/corr/abs-2106-04560}
X.~Zhai, A.~Kolesnikov, N.~Houlsby, and L.~Beyer, ``Scaling vision
  transformers,'' \emph{CoRR}, vol. abs/2106.04560, 2021.

\bibitem{gulati2020conformer}
A.~Gulati, J.~Qin, C.-C. Chiu, N.~Parmar, Y.~Zhang, J.~Yu, W.~Han, S.~Wang,
  Z.~Zhang, Y.~Wu \emph{et~al.}, ``Conformer: Convolution-augmented transformer
  for speech recognition,'' \emph{arXiv preprint arXiv:2005.08100}, 2020.

\bibitem{pan2021contrastive}
X.~Pan, L.~Wu, M.~Wang, and L.~Li, ``Contrastive learning for many-to-many
  multilingual neural machine translation,'' in \emph{Proc. of ACL}, 2021.

\bibitem{dosovitskiy2021image}
A.~Dosovitskiy, L.~Beyer, A.~Kolesnikov, D.~Weissenborn, X.~Zhai,
  T.~Unterthiner, M.~Dehghani, M.~Minderer, G.~Heigold, S.~Gelly, J.~Uszkoreit,
  and N.~Houlsby, ``An image is worth 16x16 words: Transformers for image
  recognition at scale,'' in \emph{Proc. of ICLR}, 2021.

\bibitem{brown2020language}
T.~B. Brown, B.~Mann, N.~Ryder, M.~Subbiah, J.~Kaplan, P.~Dhariwal,
  A.~Neelakantan, P.~Shyam, G.~Sastry, A.~Askell, S.~Agarwal,
  A.~Herbert{-}Voss, G.~Krueger, T.~Henighan, R.~Child, A.~Ramesh, D.~M.
  Ziegler, J.~Wu, C.~Winter, C.~Hesse, M.~Chen, E.~Sigler, M.~Litwin, S.~Gray,
  B.~Chess, J.~Clark, C.~Berner, S.~McCandlish, A.~Radford, I.~Sutskever, and
  D.~Amodei, ``Language models are few-shot learners,'' in \emph{Proc. of
  NeurIPS}, H.~Larochelle, M.~Ranzato, R.~Hadsell, M.~Balcan, and H.~Lin, Eds.,
  2020.

\bibitem{DBLP:conf/naacl/WangXWWL21}
X.~Wang, Y.~Xiong, Y.~Wei, M.~Wang, and L.~Li, ``{L}ight{S}eq: A high
  performance inference library for transformers,'' in \emph{Proc. of
  NAACL-HLT: Industry Papers}, 2021, pp. 113--120.

\bibitem{DBLP:conf/ppopp/FangYZZ21}
J.~Fang, Y.~Yu, C.~Zhao, and J.~Zhou, ``Turbotransformers: an efficient {GPU}
  serving system for transformer models,'' in \emph{Proc. of PPoPP}, J.~Lee and
  E.~Petrank, Eds., 2021, pp. 389--402.

\bibitem{DBLP:conf/sc/RajbhandariRRH20}
S.~Rajbhandari, J.~Rasley, O.~Ruwase, and Y.~He, ``Zero: memory optimizations
  toward training trillion parameter models,'' in \emph{Proc. of SC},
  C.~Cuicchi, I.~Qualters, and W.~T. Kramer, Eds., 2020, p.~20.

\bibitem{DBLP:conf/osdi/ChenMJZYSCWHCGK18}
T.~Chen, T.~Moreau, Z.~Jiang, L.~Zheng, E.~Q. Yan, H.~Shen, M.~Cowan, L.~Wang,
  Y.~Hu, L.~Ceze, C.~Guestrin, and A.~Krishnamurthy, ``{TVM:} an automated
  end-to-end optimizing compiler for deep learning,'' in \emph{PRoc. of OSDI},
  A.~C. Arpaci{-}Dusseau and G.~Voelker, Eds., 2018, pp. 578--594.

\bibitem{DBLP:conf/cvpr/JacobKCZTHAK18}
B.~Jacob, S.~Kligys, B.~Chen, M.~Zhu, M.~Tang, A.~G. Howard, H.~Adam, and
  D.~Kalenichenko, ``Quantization and training of neural networks for efficient
  integer-arithmetic-only inference,'' in \emph{Proc. of CVPR}, 2018, pp.
  2704--2713.

\bibitem{patarasuk2009bandwidth}
P.~Patarasuk and X.~Yuan, ``Bandwidth optimal all-reduce algorithms for
  clusters of workstations,'' \emph{Journal of Parallel and Distributed
  Computing}, vol.~69, no.~2, pp. 117--124, 2009.

\bibitem{li2014scaling}
M.~Li, D.~G. Andersen, J.~W. Park, A.~J. Smola, A.~Ahmed, V.~Josifovski,
  J.~Long, E.~J. Shekita, and B.-Y. Su, ``Scaling distributed machine learning
  with the parameter server,'' in \emph{11th $\{$USENIX$\}$ Symposium on
  Operating Systems Design and Implementation ($\{$OSDI$\}$ 14)}, 2014, pp.
  583--598.

\bibitem{DBLP:journals/pvldb/LiZVSNLPSVDC20}
S.~Li, Y.~Zhao, R.~Varma, O.~Salpekar, P.~Noordhuis, T.~Li, A.~Paszke,
  J.~Smith, B.~Vaughan, P.~Damania, and S.~Chintala, ``Pytorch distributed:
  Experiences on accelerating data parallel training,'' \emph{Proc. {VLDB}
  Endow.}, vol.~13, no.~12, pp. 3005--3018, 2020.

\bibitem{DBLP:conf/nips/VyasKF20}
A.~Vyas, A.~Katharopoulos, and F.~Fleuret, ``Fast transformers with clustered
  attention,'' in \emph{Proc. of NeurIPS}, H.~Larochelle, M.~Ranzato,
  R.~Hadsell, M.~Balcan, and H.~Lin, Eds., 2020.

\bibitem{DBLP:journals/corr/abs-2102-12702}
C.~Ying, G.~Ke, D.~He, and T.~Liu, ``Lazyformer: Self attention with lazy
  update,'' \emph{CoRR}, vol. abs/2102.12702, 2021.

\bibitem{DBLP:journals/corr/abs-2006-04768}
S.~Wang, B.~Z. Li, M.~Khabsa, H.~Fang, and H.~Ma, ``Linformer: Self-attention
  with linear complexity,'' \emph{CoRR}, vol. abs/2006.04768, 2020.

\bibitem{DBLP:conf/iclr/FanGJ20}
A.~Fan, E.~Grave, and A.~Joulin, ``Reducing transformer depth on demand with
  structured dropout,'' in \emph{Proc. of ICLR}, 2020.

\bibitem{DBLP:conf/nips/ZhangH20}
M.~Zhang and Y.~He, ``Accelerating training of transformer-based language
  models with progressive layer dropping,'' in \emph{Proc. of NeurIPS},
  H.~Larochelle, M.~Ranzato, R.~Hadsell, M.~Balcan, and H.~Lin, Eds., 2020.

\bibitem{peng2021random}
H.~Peng, N.~Pappas, D.~Yogatama, R.~Schwartz, N.~Smith, and L.~Kong, ``Random
  feature attention,'' in \emph{Proc. of ICLR}, 2021.

\bibitem{choromanski2021rethinking}
K.~Choromanski, V.~Likhosherstov, D.~Dohan, X.~Song, A.~Gane, T.~Sarl{\'o}s,
  P.~Hawkins, J.~Davis, A.~Mohiuddin, L.~Kaiser, D.~Belanger, L.~J. Colwell,
  and A.~Weller, ``Rethinking attention with performers,'' in \emph{Proc. of
  ICLR}, 2021.

\bibitem{DBLP:conf/iclr/LiuJHCLG020}
L.~Liu, H.~Jiang, P.~He, W.~Chen, X.~Liu, J.~Gao, and J.~Han, ``On the variance
  of the adaptive learning rate and beyond,'' in \emph{Proc. of ICLR}, 2020.

\bibitem{pmlr-v97-gong19a}
L.~Gong, D.~He, Z.~Li, T.~Qin, L.~Wang, and T.~Liu, ``Efficient training of
  {BERT} by progressively stacking,'' in \emph{Proc. of ICML}, ser. Proceedings
  of Machine Learning Research, K.~Chaudhuri and R.~Salakhutdinov, Eds.,
  vol.~97, 2019, pp. 2337--2346.

\bibitem{DBLP:conf/emnlp/LiWLJDXWZ20}
B.~Li, Z.~Wang, H.~Liu, Y.~Jiang, Q.~Du, T.~Xiao, H.~Wang, and J.~Zhu,
  ``Shallow-to-deep training for neural machine translation,'' in \emph{Proc.
  of EMNLP}, 2020, pp. 995--1005.

\bibitem{DBLP:journals/corr/abs-1904-00962}
Y.~You, J.~Li, J.~Hseu, X.~Song, J.~Demmel, and C.~Hsieh, ``Reducing {BERT}
  pre-training time from 3 days to 76 minutes,'' \emph{CoRR}, vol.
  abs/1904.00962, 2019.

\bibitem{DBLP:journals/corr/abs-2006-00719}
Z.~Yao, A.~Gholami, S.~Shen, K.~Keutzer, and M.~W. Mahoney, ``{ADAHESSIAN:} an
  adaptive second order optimizer for machine learning,'' \emph{CoRR}, vol.
  abs/2006.00719, 2020.

\bibitem{DBLP:conf/cvpr/ZhangLZLHZGGDZC20}
X.~Zhang, S.~Liu, R.~Zhang, C.~Liu, D.~Huang, S.~Zhou, J.~Guo, Q.~Guo, Z.~Du,
  T.~Zhi, and Y.~Chen, ``Fixed-point back-propagation training,'' in
  \emph{Proc. of CVPR}, 2020, pp. 2327--2335.

\bibitem{NEURIPS2019_65fc9fb4}
X.~Sun, J.~Choi, C.~Chen, N.~Wang, S.~Venkataramani, V.~Srinivasan, X.~Cui,
  W.~Zhang, and K.~Gopalakrishnan, ``Hybrid 8-bit floating point {(HFP8)}
  training and inference for deep neural networks,'' in \emph{Proc. of
  NeurIPS}, H.~M. Wallach, H.~Larochelle, A.~Beygelzimer,
  F.~d'Alch{\'{e}}{-}Buc, E.~B. Fox, and R.~Garnett, Eds., 2019, pp.
  4901--4910.

\bibitem{NEURIPS2018_335d3d1c}
N.~Wang, J.~Choi, D.~Brand, C.~Chen, and K.~Gopalakrishnan, ``Training deep
  neural networks with 8-bit floating point numbers,'' in \emph{Proc. of
  NeurIPS}, S.~Bengio, H.~M. Wallach, H.~Larochelle, K.~Grauman,
  N.~Cesa{-}Bianchi, and R.~Garnett, Eds., 2018, pp. 7686--7695.

\bibitem{DBLP:conf/aaai/LiWLDXZZ21}
B.~Li, Z.~Wang, H.~Liu, Q.~Du, T.~Xiao, C.~Zhang, and J.~Zhu, ``Learning
  light-weight translation models from deep transformer,'' in \emph{Proc. of
  AAAI}, 2021, pp. 13\,217--13\,225.

\bibitem{DBLP:conf/icml/LiWSLKKG20}
Z.~Li, E.~Wallace, S.~Shen, K.~Lin, K.~Keutzer, D.~Klein, and J.~Gonzalez,
  ``Train big, then compress: Rethinking model size for efficient training and
  inference of transformers,'' in \emph{Proc. of ICML}, ser. Proceedings of
  Machine Learning Research, vol. 119, 2020, pp. 5958--5968.

\bibitem{rasley2020deepspeed}
J.~Rasley, S.~Rajbhandari, O.~Ruwase, and Y.~He, ``Deepspeed: System
  optimizations enable training deep learning models with over 100 billion
  parameters,'' in \emph{Proc. of KDD}, R.~Gupta, Y.~Liu, J.~Tang, and B.~A.
  Prakash, Eds., 2020, pp. 3505--3506.

\bibitem{mixed-precision-training}
P.~Micikevicius, S.~Narang, J.~Alben, G.~F. Diamos, E.~Elsen, D.~Garc{\'{\i}}a,
  B.~Ginsburg, M.~Houston, O.~Kuchaiev, G.~Venkatesh, and H.~Wu, ``Mixed
  precision training,'' in \emph{Proc. of ICLR}, 2018.

\bibitem{bojar-EtAl:2014:W14-33}
\BIBentryALTinterwordspacing
O.~Bojar, C.~Buck, C.~Federmann, B.~Haddow, P.~Koehn, J.~Leveling, C.~Monz,
  P.~Pecina, M.~Post, H.~Saint-Amand, R.~Soricut, L.~Specia, and A.~Tamchyna,
  ``Findings of the 2014 workshop on statistical machine translation,'' in
  \emph{Proceedings of the Ninth Workshop on Statistical Machine
  Translation}.\hskip 1em plus 0.5em minus 0.4em\relax Baltimore, Maryland,
  USA: Association for Computational Linguistics, June 2014, pp. 12--58.
  [Online]. Available: \url{http://www.aclweb.org/anthology/W/W14/W14-3302}
\BIBentrySTDinterwordspacing

\bibitem{Krizhevsky09learningmultiple}
A.~Krizhevsky, ``Learning multiple layers of features from tiny images,'' Tech.
  Rep., 2009.

\bibitem{DBLP:conf/iclr/WangSMHLB19}
\BIBentryALTinterwordspacing
A.~Wang, A.~Singh, J.~Michael, F.~Hill, O.~Levy, and S.~R. Bowman, ``{GLUE:}
  {A} multi-task benchmark and analysis platform for natural language
  understanding,'' in \emph{7th International Conference on Learning
  Representations, {ICLR} 2019, New Orleans, LA, USA, May 6-9, 2019}.\hskip 1em
  plus 0.5em minus 0.4em\relax OpenReview.net, 2019. [Online]. Available:
  \url{https://openreview.net/forum?id=rJ4km2R5t7}
\BIBentrySTDinterwordspacing

\bibitem{DBLP:conf/iclr/MerityX0S17}
\BIBentryALTinterwordspacing
S.~Merity, C.~Xiong, J.~Bradbury, and R.~Socher, ``Pointer sentinel mixture
  models,'' in \emph{5th International Conference on Learning Representations,
  {ICLR} 2017, Toulon, France, April 24-26, 2017, Conference Track
  Proceedings}.\hskip 1em plus 0.5em minus 0.4em\relax OpenReview.net, 2017.
  [Online]. Available: \url{https://openreview.net/forum?id=Byj72udxe}
\BIBentrySTDinterwordspacing

\bibitem{DBLP:conf/naacl/OttEBFGNGA19}
M.~Ott, S.~Edunov, A.~Baevski, A.~Fan, S.~Gross, N.~Ng, D.~Grangier, and
  M.~Auli, ``fairseq: A fast, extensible toolkit for sequence modeling,'' in
  \emph{Proc. of NAACL-Demonstrations}, 2019, pp. 48--53.

\bibitem{wikitext}
S.~Merity, C.~Xiong, J.~Bradbury, and R.~Socher, ``Pointer sentinel mixture
  models,'' 2016.

\end{thebibliography}

\end{document}